\documentclass[11pt]{article}

\usepackage{hyperref}
\usepackage{fullpage}
\usepackage{amsmath,amsfonts,amsthm,amssymb,xspace,bm, verbatim,dsfont,mathtools}
\usepackage{natbib}
\usepackage{graphicx}
\usepackage{url,algorithm,algorithmic}
\usepackage{color}
\usepackage{epstopdf}
\usepackage{appendix}
\theoremstyle{plain}
\newtheorem{theorem}{Theorem}[section]
\newtheorem{lemma}[theorem]{Lemma}
\newtheorem*{lemma*}{Lemma}
\newtheorem{claim}[theorem]{Claim}
\newtheorem{corollary}[theorem]{Corollary}
\newtheorem{definition}[theorem]{Definition}
\theoremstyle{definition}
\newcommand{\vecfont}[1]{\mathbf{#1}}
\newcommand{\mat}[1]{\mathbf{#1}}

\newcommand{\e}{\vecfont{e}}

\newcommand{\note}[1]{\marginpar{\tiny *note in TeX*}}
\newcommand{\ignore}[1]{}

\renewcommand{\phi}{\varphi}

\newcommand{\R}{\mathbb{R}}

\newcommand{\ip}[2]{\langle #1, #2 \rangle}

\def\Po{P_{\Omega}}
\def\fnd{\frac{n}{d}}
\def\lv{\left\vert}
\def\rv{\right\vert}
\def\lV{\left\Vert}
\def\rV{\right\Vert}

\begin{document}

% If your paper is accepted and the title of your paper is very long,
% the style will print as headings an error message. Use the following
% command to supply a shorter title of your paper so that it can be
% used as headings.
%
%\runningtitle{I use this title instead because the last one was very long}

% If your paper is accepted and the number of authors is large, the
% style will print as headings an error message. Use the following
% command to supply a shorter version of the authors names so that
% they can be used as headings (for example, use only the surnames)
%
%\runningauthor{Surname 1, Surname 2, Surname 3, ...., Surname n}

%\twocolumn[
\title{Universal Matrix Completion}
\author{
Srinadh Bhojanapalli\\
{The University of Texas at Austin}\\
{bsrinadh@utexas.edu}
\and
Prateek Jain\\
{Microsoft Research, India}\\
{prajain@microsoft.com}
}
\maketitle

\begin{abstract}
The problem of low-rank matrix completion has recently generated a lot of interest leading to several results that offer exact solutions to the problem.  
%Recently, several methods have been shown to solve the problem of low-rank matrix completion exactly. %Several recent works provide provabshow that the low-rank matrix completion problem can be solved exactly. 
However, in order to do so, these methods make assumptions that can be quite restrictive in practice. More specifically, the methods assume that: a) the observed indices are sampled uniformly at random, and b) for every new matrix, the observed indices are sampled \emph{afresh}. In this work, we address these issues by providing a universal recovery guarantee for matrix completion that works for a variety of sampling schemes. In particular, we show that if the set of sampled indices come from the edges of a bipartite graph with large spectral gap (i.e. gap between the first and the second singular value), then the nuclear norm minimization based method exactly recovers all low-rank matrices that satisfy certain incoherence properties. 
%However, most of the existing methods require the following  assumptions that are rather restrictive: a) the observed entries are sampled uniformly at random, and b) for every new matrix, the indices of observed entries need to be sampled {\em afresh}. In this work, we address these issues by providing a {\em universal} recovery guarantee for matrix completion that works for a variety of sampling schemes. In particular, we show that if the set of sampled indices ($\Omega$) come from the edges of a bipartite graph with large spectral gap (i.e. gap between the first and the second singular values), then the nuclear-norm minimization based method exactly recovers {\em all} low-rank matrices that satisfy certain incoherence properties. %Our generic result guarantees {\em universal} recovery by nuclear-norm based method as long as the sampling graph has certain spectral gap. 
Moreover, we also show that under certain stricter incoherence conditions, $O(nr^2)$ uniformly sampled entries are enough to recover any rank-$r$ $n\times n$ matrix, in contrast to the $O(nr\log n)$ sample complexity required by other matrix completion algorithms as well as existing analyses of the nuclear norm method.%Moreover, our result shows that, under certain stricter incoherence conditions,  $O(nr^2)$ uniformly sampled entries are enough to recover any rank-$r$ matrix, in contrast to $O(n r \log n)$ entries required by the existing analysis of the nuclear-norm method;  other matrix completion algorithms are also known to require similar $O(nr \log n)$ sample complexity. % (or by any other matrix completion algorithm in fact). %, unlike existing results that require each index to be sampled uniformly.
\end{abstract}

\section{Introduction}
In this paper, we study the problem of {\em universal} low-rank matrix completion. Low-rank matrix completion is an important problem with several applications in areas such as recommendation systems, sketching, and quantum tomography  \citep{recht2010guaranteed, candes2009exact, gross2010quantum}. The goal in matrix completion is to recover a rank-$r$ matrix, given a small number of entries of the matrix. That is, to find a matrix $M$ given values $\{ M_{ij}, (i,j)\in \Omega \}$, where $\Omega$ is the set of observed indices. 

Recently, several  methods with provable guarantees have been proposed for solving the problem under the following two assumptions: a) $M$ is incoherent, b) $\Omega$ is sampled {\em uniformly} and $|\Omega|\geq c n r \log n$. Moreover, $\Omega$ needs to be re-sampled for each matrix $M$ that is to be recovered, i.e., the same $\Omega$ cannot be re-used without worsening the guarantees significantly. 

While the first assumption can be shown to be necessary for any matrix oblivious sampling, the second assumption is relatively restrictive and might not hold in several practical settings. The main goal of this work is to develop a general result that can handle other sampling schemes as well. 
Moreover, we aim to develop a universal method where one fixed $\Omega$ would be enough to recover any low-rank matrix $M$. Such a universal recovery result is highly desirable in several signal processing applications, where the goal is to design one $\Omega$ that can recover any low-rank signal matrix $M$ by observing $M$ over $\Omega$ alone. 

To this end, we reduce the problem of recoverability using an index set $\Omega$ to the spectral gap (gap between largest and second largest singular values) of $\mathcal{G}$, where $\mathcal{G}$ is a bipartite graph whose biadjacency matrix $G\in \R^{n\times n}$ is given by: $G_{ij}=1 ~ \text{iff} ~ (i,j)\in \Omega$ and $G_{ij}=0$ otherwise. %Consider a  
In particular, we show that if $\mathcal{G}$ has a large enough spectral gap and if the rank-$r$ matrix $M$ satisfies the standard incoherence property, then the best  rank-$r$ approximation of $P_\Omega(M)$ (see \eqref{eq:pomega}) itself is enough to get a ``reasonable'' approximation to $M$ (see Theorem~\ref{thm:init} for details). 

Note that our approximation result is similar to Theorem 1.1 of \citep{keshavan2010matrix}, but our result holds for any $\Omega$ with large spectral gap unlike \citep{keshavan2010matrix} that requires uniform sampling. On the other hand, we require explicit incoherence condition on singular vectors of $M$, while the result of \citep{keshavan2010matrix} only requires a bound on $M_{max}=\max_{ij}M_{ij}$. The later assumption is strictly weaker assumption; the assumptions coincide for PSD matrices. 
%Furthermore,  the result of \citep{keshavan2010matrix} is a direct corollary of our result when $\Omega$ is sampled uniformly. Finally, our proof brings out spectral gap as the key factor in matrix approximation and  provides a significantly simpler proof compared to that of \citep{keshavan2010matrix} (our proof  fits in half a column). 

Next, we show that by assuming certain stronger incoherence properties,  the number of samples required by the popular nuclear-norm  minimization method \citep{candes2009exact} to recover back $M$ depends {\em only} on $n, r$ and the spectral gap of the $d$-regular bipartite graph $\mathcal{G}$. %, i.e., $gap_G=\lambda_1(G)-\lambda_2(G)$ where $\lambda_i$ is the $i$-th largest singular value of $G$. 
In particular, we require $d\geq \sigma_2(G)\cdot r$, where $\sigma_2(G)$ is the second largest singular value of $G$. Hence, if $\sigma_2(G)=O(\sqrt{d})$, i.e., if $\mathcal{G}$ is an expander, then $|\Omega|=nd =O(n r^2)$ samples suffice for exact recovery. %=\max\{n_1, n_2 \} d =O(\max\{n_1, n_2 \}r^2)$ samples suffice for exact recovery.  %the number of samples required are $O(\max\{n_1, n_2 \} \frac{\sigma_2(G)^2}{ )$. Note that nuclear norm of a matrix is defined as the sum of singular values of the matrix.  

Our recovery results applies to {\em any} low rank matrix $M$ that satisfies the stronger incoherence property, given that the fixed graph $\mathcal{G}$ has a large spectral gap. %, and hence provides a universal guarantee for matrix completion; 
To the best of our knowledge, this is the {\it first universal guarantee for matrix completion}. Furthermore, using recent results by \citep{feige2005spectral} we show that for the standard uniform sampling of $\Omega$, %a regular Erd\"os-R\'enyi graph, which  corresponds to uniform sampling of $\Omega$, 
only $O(n r^2)$ samples suffice for exact recovery of a rank-$r$ matrix $M$ that satisfies a stronger incoherence condition (see  $A2$  in Section~\ref{sec:formulation}). %Also, our corresponding result for approximate recover of $M$ (see Theorem~\ref{thm:init}) matches the result by \citep{keshavan2010matrix} but our proof is significantly simpler (just 4 lines) than that of~\citep{keshavan2010matrix}. 

Next, we discuss the stronger incoherence property that we require for our universal recovery guarantees. In particular, we show that the standard incoherence condition alone cannot provide universal recovery with any graph $\mathcal{G}$ and hence a stronger incoherence property is required. 

Finally, we empirically demonstrate our observation that, instead of the number of samples, the spectral gap of $\mathcal{G}$ is what really governs recoverability of the true matrix. In particular, we construct a family of graphs based on the stochastic block model and show that the probability of success grows linearly with the spectral gap, irrespective of the number of samples. 

{\bf Notation}: We denote matrices by capital letters (e.g. $U$) and vectors by small letters (e.g. $u$). $U^T$ denotes the transpose of the matrix $U$. $U_{ij}$ represents the $(i,j)$-th element of $U$. $U_i$ represents the $i$-th column of $U$ and $U^i$ represents the $i$-th row of $U$ (but in column format). $\|u\|$ represents the $L_2$ norm of $u$ and $\|U\|$ represents the spectral norm of $U$, i.e., $\|U\|=\max_{x: \|x\| \leq1} \|Ux\|$. $\|U\|_F$ represents the Frobenius norm and $\|U\|_{\infty}$ is the absolute maximum element of $U$. $C=A.B$ represents the Hadamard product of $A, B$, i.e., $C_{ij}=A_{ij}B_{ij}$. Similarly, $(u.v)_i=u_iv_i$. $\mat{1}$ denotes the all $1$'s vector. $\mat{1}_{\perp}$ represents a {\em unit} vector that is perpendicular to $\mat{1}$ and is  determined appropriately by the context. %with some vector that is that is determined appropriately by the context. 

{\bf Paper Organization}: In the next section, we discuss some  related works. Then, in Section~\ref{sec:formulation}, we define the problem of matrix completion and the bipartite graph $\mathcal{G}$ that we use. We present our main results in Section~\ref{sec:results} and discuss the additional incoherence assumption in Section~\ref{sec:conds}. In Section~\ref{sec:simulations}, we present  observations from our empirical study. Finally in Section~\ref{sec:proof_sketch}, we provide the proof of our exact recovery result. %a high-level proof-sketch of our exact recovery result. %discuss the proofs of some of our main results.

%%% Local Variables: 
%%% mode: latex
%%% TeX-master: "deterministic_MC"
%%% End: 

\section{Related work}\label{sec:related}
{\bf Matrix completion}: In a seminal paper on matrix completion, \citep{candes2009exact} showed that any $n \times n$ incoherent matrix of rank $r$ can be recovered from $C n^{1.2}r \log(n)$ uniform random samples using nuclear norm minimization. %Nuclear norm of a matrix is defined as the sum of singular values of the matrix. 
Later, assuming the matrix to be {\it strongly incoherent}, \citep{candes2010power} improved the sample complexity for nuclear norm minimization method to  ${O}(nr\log^6(n))$. Subsequently, \citep{recht2009simpler, gross2011recovering} generalized this result for any incoherent matrix using matrix Bernstein inequalities and presented significantly simpler proofs.  Concurrently other algorithms were shown to recover incoherent matrices using ${O}(nr\log(n))$ (or worse) samples such as: SVD followed by descent on Grassmanian manifold \citep{keshavan2010matrix}, alternating minimization \citep{jain2012low}. %\citep{chandrasekaran2011rank, candes2011robust} have given algorithms with exact recovery guarantees for a matrix when the samples are corrupted by sparse errors and noise~\citep{candes2010matrix}. 
We note that all  the above mentioned results need to assume a rather restrictive sampling scheme, i.e., each entry is sampled uniformly at random and furthermore require a fresh set of samples for each new matrix. Moreover, the number of samples required is at least $O(nr \log n)$. 

{\bf Other sampling schemes}: Recently, there has been some results for different type of sampling schemes such as power-law distributions \citep{meka2009matrix}, but here again universal results are not known and furthermore the proposed algorithms are not robust to noise. Another line of work has been to devise sampling schemes {\em dependent} on the data matrix \citep{chen2013coherent}, \citep{KiralyT12}. Naturally, these schemes cannot be universal as the sampling scheme itself is dependent on the data matrix. Furthermore, practicality of such schemes is not clear  a priori. %Recently in \citep{chen2013coherent} it is shown that even a coherent matrix can be recovered in $\mathcal{O}(nr\log^2(n))$ if the sampling is proportional to the local coherences of the matrix.

{\bf Universality:} Universality is an important property in signal processing or sketching applications, as the goal there is to have one fixed sampling operator that performs well for all the given signals. While, universality results are well known for several other sensing problems, such as sparse vector recovery~\citep{candes2005decoding}, one-bit compressive sensing~\citep{gopi2013one}, similar results for low-rank matrix sensing are mostly restricted to RIP-type operators~\citep{recht2010guaranteed, liu2011universal}. Unfortunately, RIP-type of operators are typically dense, requiring knowledge of all elements of matrix to get observations, and also require large storage/computational complexity. On the other hand sampling individual elements is a sparse operator and hence computationally efficient. Hence, for several signal processing applications, universal matrix completion results are critical. %there is a need to study universal matrix completion type operators that are known to be sparse and are computationally efficient.% are very useful for such applications. %Universality of the matrix completion operators is an open problem that has several 

In fact, several recent works have studied problems similar to that of universal matrix completion. For example, \citep{KiralyT12, heiman2013deterministic} and~\citep{lee2013matrix}. However, there are critical differences in our results/approaches that we now highlight. In~\citep{KiralyT12} authors consider an algebraic approach to analyze sufficient conditions for matrix completion. %However their results do handle all the matrices and does not apply to measure-zero matrices which is a notion that they introduce.
While they propose interesting deterministic sufficient conditions, the algorithm analyzed in the paper requires solving an NP hard problem. In contrast we analyze the nuclear norm minimization method which is known to have several efficient implementations. In~\citep{heiman2013deterministic} and~\citep{lee2013matrix} authors consider sampling based on expanders but only provide generalization error bounds rather than exact recovery guarantee. Moreover, the  recovered matrix using their algorithm need not have a low-rank.

%%% Local Variables: 
%%% mode: latex
%%% TeX-master: "deterministic_MC"
%%% End: 

\section{Problem Definition \& Assumptions}\label{sec:formulation}
Let $M\in \R^{n_1 \times n_2}$ be a rank-$r$ matrix and let $n_1 \geq n_2$. Define $n=\max\{n_1, n_2\}=n_1$. Let $M=U\Sigma V^T$ be the SVD of $M$ and let $\sigma_1\geq \sigma_2\dots\geq \sigma_r$ be the singular values of $M$. We observe a small number of entries of $M$ indexed by a set $\Omega\in [n_1]\times [n_2]$. That is, we observe $M_{ij}, \forall (i,j) \in \Omega$. Define the sampling operator $P_\Omega: \R^{n_1\times n_2}\rightarrow \R^{n_1\times n_2}$ as: 
\begin{equation}
  \label{eq:pomega}
  P_{\Omega}(M)=\begin{cases}M_{ij},&\ \text{ if }(i,j) \in \Omega, \\0,&\ \text{ if } (i,j) \not\in \Omega. \end{cases}
\end{equation}
Next, we define a bipartite graph associated with the sampling operator $P_\Omega$. That is, let $\mathcal{G}=(V, E)$ be a bipartite graph where $V=\{1,2,\dots,n_1\}\cup \{1,2,\dots, n_2\}$ and $(i,j) \in E\text{ iff }(i,j)\in \Omega$. Let $G\in \R^{n_1\times n_2}$ be the biadjacency matrix of the bipartite graph $\mathcal{G}$ with $G_{ij}=1\text{ iff }(i,j) \in \Omega$. Note that, $P_\Omega(M)=M. G$, where $.$ denotes the Hadamard  product. 

Now, the goal in {\em universal} matrix completion is to design a set $\Omega$ and a recovery algorithm, s.t., all rank-$r$ matrices $M$ can be recovered using only $P_\Omega(M)$. In the next section,  we  present two results for this problem. Our first result gives an approximate solution to the matrix completion problem and our second result gives exact recovery guarantees. 

For our results, we require $\mathcal{G}$, that is associated with $\Omega$, to be a $d$-regular bipartite graph with large spectral gap. More concretely, we require the following two properties from the sampling graph $\mathcal{G}$: 
%$nd$ samples of matrix $M$ according to the deterministic sampling operator $P_{\Omega}$ that operates on matrices in $\R^{n_1 \times n_2}$, where $\Omega$ is a subset of $[n_1] \times [n_2]$. Size of $\Omega, |\Omega|=nd$.  \[ \Po(M)_{ij}= M_{ij} \mbox{ if } (i, j) \in \Omega, 0 \; \mbox{else.}\]
%\noindent We can also represent $\Po(M)$ as $M.*G$ where $.*$ denotes the Hadamard product (elementwise product) and $G_{ij}=1$ \;\mbox{if} $(i, j) \in \Omega, 0 $ else. $G$ is the biadjacency matrix of a undirected bipartite graph with $n_1$ and $n_2$ nodes on each side. 

{\bf Assumptions on $G/\Omega$}: 
%We make the following assumptions on $G/\Omega$.
\begin{itemize}
\item {\bf G1} \quad  Top singular vectors of $G$ are all $1$'s vector.
\item {\bf G2} \quad $\sigma_1(G) =d$ and $\sigma_2(G) \leq C\sqrt{d}$.
\end{itemize}
Note that as the graph is $d$-regular, hence $|\Omega|=nd$. 

\noindent The eigenvalues of the adjacency matrix of the bipartite graph $\mathcal{G}$ are $\{\sigma_i(G),-\sigma_i(G)\}, i=1,..n$. We state all the definitions in terms of singular values of $G$ instead of the eigenvalues of the adjacency matrix. The above two properties are satisfied by a class of expander graphs called Ramanujan graphs; in fact, Ramanujan graphs are defined by using this spectral gap property:  %The spectral gap of the graph is $\lambda_1 -\lambda_2.$
\begin{definition}[Ramanujan graph~\citep{hlw06}]
Let $\sigma_1(G), \sigma_2(G),...,\sigma_n(G)$ be the singular values of $G$ in decreasing order. Then, a $d$-regular bipartite graph $\mathcal{G}$ is a Ramanujan graph if $\sigma_2(G) \leq 2\sqrt{d-1}$.
\end{definition}

\noindent Ramanujan graphs $\mathcal{G}$ are well-studied in literature and there exists several randomized/deterministic methods to generate such graphs. We briefly discuss a couple of popular constructions in Section~\ref{sec:nuclear}.% Also, note that as the graph is $d$-regular, hence $|\Omega|=nd$. %Generating graphs $G$ that satisfy these propre\noindent There exists graphs that satisfy these properties in expander families.

{\bf Incoherence assumptions}: Now, we present incoherence assumptions that we impose on  $M$:  %Also assume the following incoherence conditions on matrix $M$.  
\begin{align}
{\mathbf A1} ~ &||U^i||^2 \leq \frac{\mu_0 r}{n_1}, \forall i ~ ~\mbox{ and 	}~ ~   ||V^j||^2 \leq \frac{\mu_0 r}{n_2}, \forall j \label{eq:assumption1}\\
%&{\mathbf A2} \quad \frac{\mu_1 d}{n}||x||^2 \leq \sum_{k \in S} (U^{k^T}x)^2 \leq \frac{\mu_0 d} {n} ||x||^2,\label{eq:assumption2} \\
{\mathbf A2} ~ &\|\sum_{k\in S} \frac{n_1}{d} U^k U^{k^T} -I\| \leq \delta_d, \forall S\subset [n_1], |S|=d  ~ ~\mbox{ and 	}~ ~  \nonumber\\ &\|\sum_{k\in S} \frac{n_2}{d'} V^k V^{k^T} -I\| \leq \delta_d, \forall S\subset [n_2], |S|=d'. \label{eq:assumption3}
%\quad \mbox{ and} \quad  \|\sum_{k\in S} \fnd U^K U_{\perp}^{k^T}\| \leq \delta_d^{\perp}
\end{align}%for all sets $S$, $S \subset [n]$ and $|S| = d$.
$d'=d n_2/n_1$. Note that $A1$ is the standard incoherence assumption required by most of the existing matrix completion results. However, $A2$ is a stricter assumption than $A1$ and is similar to the stronger incoherence property introduced by \citep{candes2010power}. We discuss necessity of such assumption for universal matrix completion in Section~\ref{sec:conds}.

\def\Pt{\mathcal{P}_{T}}
\def\Ptp{\mathcal{P}_{T^{\perp}}}
\def\Pon{\mathcal{P}_\mat{1}}
\def\Ponp{\mathcal{P}_{\mat{1}^{\perp}}}
\section{Main Results}\label{sec:results}
We now present our main results for the matrix completion problem. We assume that $\Omega$ is generated using a bipartite $d$-regular expander and satisfies $G1$ and $G2$ (see Section~\ref{sec:formulation}). Our first result shows that, if $M$ satisfies $A1$, then the best rank-$r$ approximation of $P_\Omega(M)$ is ``close'' to $M$ and hence serves as a good approximation for $M$ that can also be used for initialization of other methods like alternating least squares. Our second results shows that if $M$ satisfies both $A1$ and $A2$, then using nuclear-norm minimization based method, $P_\Omega(M)$  can be used to recover back $M$ exactly. 
\subsection{Matrix approximation}
\begin{theorem}\label{thm:init}
Let $\mathcal{G}$ be a $d$-regular bipartite graph satisfying $G1$ and $G2$. Let $M$ be a rank-$r$ matrix that satisfies assumption $A1$. Then, %Given $\Po$ satisfying the assumptions {\bf (G1)} and {\bf (G2)} and a rank $r$ matrix $M$ satisfying assumption {\bf (A1)}, then 
\[\lV\frac{n}{d}\Po(M)-M\rV \leq  \frac{ C \mu_0 r} {\sqrt{d}} ||M||.\]\label{lemma:po_spectral}
That is, $\|\frac{n}{d}P_k(P_\Omega(M))-M\| \leq \frac{ 2C \mu_0 r} {\sqrt{d}} ||M||,$ for any $k \geq r$, where $P_k(A)$ is the best rank-$k$ approximation of $A$ and can be obtained using top-$k$ singular vectors of $A$. 
\end{theorem}
\noindent Now, if $M$ is a PSD matrix then the above result is exactly same as the Theorem 1.1 of \citep{keshavan2010matrix}. For non-PSD matrices, our result requires a bound on norm of each row of singular vectors of $M$, while the result of \citep{keshavan2010matrix} only requires a bound on the largest element of $M$, hence is similar to our requirement but is strictly weaker as well. 

On the other hand, our result holds for all $M$ for a given $\Omega$, if $\Omega$'s associated graph $\mathcal{G}$ satisfies both $G1$ and $G2$. %and holds for any $G$ (i.e., $\Omega$) that satisfies $G_1$ and $G_2$. 
Moreover, if $\mathcal{G}$ is generated using an Erdos-Renyi graph then, after a standard trimming step, the above theorem directly implies(for PSD matrices) Theorem 1.1 of \citep{keshavan2010matrix}. Finally, we would like to stress that our proof is significantly simpler and is able to exploit the fact that Erd\"os-R\'enyi graphs have good spectral gap in a fairly straightforward and intuitive manner. 

We now present a detailed proof of the above theorem. 
\begin{proof}
Let $M=U\Sigma V^T$,  $U, V \in \R^{n \times r}$. Note that, \[\|\frac{n}{d}\Po(M)-M\|=\max_{\stackrel{\{x, y: \|x\|_2=1,}{\|y\|_2=1\}}}y^T(\frac{n}{d}\Po(M)-M)x.\] 

Now, \begin{align}y^T(\frac{n}{d}\Po(M)-M)x=\sum_{i=1}^r \left(\fnd \sigma_i (y.U_i)^TG(x.V_i)- \sigma_i(y^T U_i)(x^T V_i)\right).\end{align}
Let $y.U_i=\alpha_i {\bf 1}+\beta_i {\bf 1_\perp^i}$. Then, $\alpha_i=\frac{{\bf 1}^T (y.U_i)}{n}=\frac{y^T U_i}{n}$. Hence, 
\begin{align}
  y^T(\frac{n}{d}\Po(M)-M)x  =&\sum_{i=1}^r \left( \sigma_i (y^T U_i x^T V_i + \fnd \beta_i {\bf 1_\perp^i}^T G (x.V_i))-\sigma_i y^T U_i x^T V_i \right) \nonumber\\
\stackrel{\zeta_1}{\leq}& \frac{Cn}{\sqrt{d}}\sum_i \sigma_i \beta_i \|x.V_i\|_2\stackrel{\zeta_2}{\leq} \sigma_1 \frac{Cn}{\sqrt{d}}\sqrt{\sum_i \beta_i^2} \sqrt{\sum_i \|x.V_i\|_2^2}.\label{eq:approx_1}%\nonumber\\&\leq \sigma_1 \frac{C\mu_0 r}{\sqrt{d}},
\end{align}
where $\zeta_1$ follows from assumption $G2$ and $\zeta_2$ follows from the Cauchy-Schwarz inequality. Now,%the second inequality follows from the assumption $G2$ and the last inequality follows from the following: 
\begin{align} 
\sum_{i=1}^r \beta_i^2 \leq \sum_{i=1}^r \|y.U_i\|^2 =\sum_{j=1}^n \sum_{i=1}^r  y_j^2 U_{ji}^2  \stackrel{\zeta_1}{\leq} \sum_{j=1}^n y_j^2 \frac{\mu_0 r}{n}\stackrel{\zeta_2}{=} \frac{\mu_0 r}{n},\label{eq:approx_2}
\end{align}
where $\zeta_1$ follows from $A1$ and $\zeta_2$ follows by using $\|y\|_2=1$. Using similar argument as above, $\sum_{i=1}^r \|x.V_i\|^2 \leq \frac{\mu_0 r}{n}.$
Theorem now follows by using \eqref{eq:approx_1}, \eqref{eq:approx_2}, and the above inequality. The proof of the second part is given in appendix~\ref{thm:init2}.
\end{proof}

\subsection{Nuclear norm minimization}\label{sec:nuclear}
We now present our result for exact recovery  of the matrix $M$ using  $\Po(M)$ alone. For recovery, we use the standard nuclear norm minimization algorithm, i.e., we obtain a matrix $X$ by solving the following convex optimization problem:
\begin{equation} \label{eq:nnm}
 \begin{split}
 &\min  \quad \|X\|_* \\
&\mbox{ s. t. } \Po(X)=\Po(M), 
\end{split}
\end{equation}
where $\|X\|_*$ denotes the nuclear norm of $X$; nuclear norm of $X$ is equal to the sum of its singular values. 

As mentioned in Section~\ref{sec:related}, nuclear norm minimization technique is a popular technique for the low-rank matrix completion problem and has been shown to provably  recover the true matrix, assuming that $\Omega$ is sampled uniformly at random and $|\Omega|\geq c n r \log n$ \citep{candes2010power}. 

Below, we provide a universal recovery result for the nuclear-norm minimization method as long as the samples $\Omega$ come from $G$ that satisfies $G1$ and $G2$. 
%%%%%%%%%%%%%%%%%%%%%-main theorem-%%%%%%%%%%%%%%%%%%%%%%%%
\begin{theorem}\label{thm:main}
Let $M$ be an $n_1 \times n_2$ matrix of rank $r$ satisfying assumptions { (A1)} and {(A2)} with $\delta_d \leq \frac{1}{6},$ and $\Omega$ is generated from a $d$-regular graph $\mathcal{G}$ that satisfies the assumptions (G1) and (G2). Also, let $d\geq 36C^2\mu_0^2r^2$, i.e., $|\Omega| =n d\geq  36 C^2\mu_0^2 r^2 \max\{n_1, n_2\}$. Then $M$ is the unique optimum of problem~\eqref{eq:nnm}. %That is, if , then all rank-$r$ $M$ (satisfying $A1, A2$) can be recovered 
\end{theorem}

Note that the above result requires only deterministic constraints on the sampling operator $\Po$ and guarantees exact recovery for {\em any} matrix $M$ that satisfies $A1, A2$. %is strongly incoherent, hence provides a universal recovery guarantee. 
As mentioned earlier, $A2$ is a stronger assumption than $A1$. But as we show in Section~\ref{sec:conds}, universal recovery  is not possible with  assumption $A1$ alone.% for a justification of $A2$. 

We can  use the above theorem to derive results for  several interesting sampling schemes such as random $d$-regular graphs. Using Theorem 1 in~\citep{friedman2003proof},  the second singular value of a {\em random  $d$-regular graph}  is $\leq 2\sqrt{d-1} +\epsilon$, for every $\epsilon >0$, with high probability. Hence, a random $d$-regular graph, with high probability,  obeys $G1$ and $G2$ which implies the following exact recovery result: %For a more detailed discussion on graphs with high spectral gap look in section~\ref{sec:expanders}

\begin{corollary}\label{cor:d_regular}
Let $M$ be an $n_1 \times n_2$ matrix of rank $r$ satisfying assumptions { (A1)} and {(A2)} with $\delta_d \leq \frac{1}{6},$ and $\Omega$ is generated from a random $d$-regular graph, then $M$ is the unique optimum of program~\eqref{eq:nnm} when  $d \geq  36*4 \mu_0^2 r^2$, with high probability.
\end{corollary}
Note that the standard completion results such as~\citep{candes2009exact}, \citep{keshavan2010matrix} generate $\Omega$ using Erd\"os-R\'enyi graph, which are slightly different than the random $d$-regular graph we considered above. However, \citep{feige2005spectral} showed that the second largest singular value  of the Erd\"os-R\'enyi graph, $\mathcal{G}(n_1, n_2, p)$, is ${O}(\sqrt{d})$ when $p$ is ${\Theta}(\log(n_1)/n_2)$. Interestingly, even when $p=c/n_2$, i.e. $n_2\cdot p$ is a constant, trimming the graph (i.e., removing few nodes with high degree) gives a graph $G$ s.t. $\sigma_2(G)={O}(\sqrt{d})$. Hence, we can again apply Theorem~\ref{thm:main} to obtain the following result:%In \citep{keshavan2010matrix} similar trimming of the input samples is done to get good approximation of the sampled matrix.
\begin{corollary}\label{cor:gnp}
Let $M$ be an $n_1 \times n_2$ matrix of rank $r$ satisfying assumptions { (A1)} and {(A2)} with $\delta_d \leq \frac{1}{6},$ and $\Omega$ is generated from a $\mathcal{G}(n, p)$ graph after trimming, then $M$ is the unique optimum of program~\eqref{eq:nnm} when $p \geq  \frac{36 c \mu_0^2 r^2}{\min\{n_1, n_2\}} $, with high probability.
\end{corollary}\vspace*{-5pt}
While the above two results exploit the fact that a random graph is almost a Ramanujan expander and hence our general recovery result can be applied, the graph construction is still randomized. Interestingly,  %There exists several Ramanujan expanders which uses significantly less randomization than Erd\"os-R\'enyi type of expanders and some constructions are in fact completely deterministic. See 
\citep{ lubotzky1988ramanujan, margulis1988explicit, morgenstern1994existence} proposed explicit deterministic constructions of Ramanujan graphs when $d-1$ is a prime power. Moreover, \citep{marcus2013interlacing} showed that bipartite Ramanujan graphs exist for all $n$ and $d$. However, explicit construction for all $n$ and $d$ still remains an open problem.

\section{Discussion}\label{sec:conds}
In this section, we discuss the two assumptions $A1$ and $A2$ that are mentioned in Section~\ref{sec:formulation}. %that we impose on the underlying matrix $M$ that is to be recovered in an universal method. 

Note that $A1$ is a standard assumption that is used by most of the existing approaches \citep{candes2009exact}, \citep{keshavan2010matrix}. Moreover, it is easy to show that for any matrix ``oblivious'' sampling  approach, this assumption is necessarily required for exact recovery. For example, if $G_{ij}=0$, i.e., $(i,j)$-th element is not observed then we cannot recover $M=\e_i \e_j^T$. 

However, $A2$ is a slightly non-standard assumption and intuitively it requires the singular vectors of $M$ to satisfy RIP. Note that $A2$ is similar to the strong incoherence property introduced by \citep{candes2010power}. Below we show the connection between strong incoherence property (SIP) assumed in \citep{candes2010power} and assumption $A2$.% (see Lemma~\ref{lem:sip}). 

\begin{claim}\label{cl:sip}
  Let $M\in \R^{n_1\times n_2}$ be a rank-$r$ matrix. Let $M=U\Sigma V^T$ satisfy SIP i.e., 
  \begin{align}
  \hspace*{-6pt} |\langle \e_i, UU^T \e_j\rangle - \frac{r}{n_1} {\bm 1}_{i=j}|\leq \mu_1 \frac{\sqrt{r}}{n_1}, \forall 1\leq i,j \leq n_1, \nonumber\\
\hspace*{-6pt}|\langle \e_i, VV^T \e_j\rangle - \frac{r}{n_1} {\bm 1}_{i=j}|\leq \mu_1 \frac{\sqrt{r}}{n_2}, \forall 1\leq i,j \leq n_2. 
    \label{eq:sip}
  \end{align}
Then, $M$ satisfies $A2$ for all $d\geq r$ and $\delta_{d} \leq {\mu_1}{\sqrt{r}}$. 
\end{claim}
Note that the above claim holds with $\delta_d=\mu_1\sqrt{r},\ \forall d\geq r$. This bound is independent of $d$ and hence weak; as $d$ becomes close to $n_1$, $\lV \frac{n_1}{d}\sum_{k\in S} U^k{U^k}^T-I\rV$ gets close to 0 since $U^T U= I$. We leave the task of obtaining a stronger bound as an open problem.
%\begin{proof}
%  Note that for hermitian matrix $A\in \R^{r\times r}$: $\|A\| \leq \max_{i \in [r]} \sum_{j}|A_{ij}|$. Hence, for any set $S$, s.t. $|S|=r$, we have: 
%\begin{align*}&\hspace*{-15pt}\lV \frac{n_1}{r}\sum_{k\in S} U^k{U^k}^T-I\rV \leq \max_{i \in [r]} \sum_j \left| \frac{n_1}{r}\sum_{k=1}^d  U_{ki}U_{kj}-1_{i=j}\right|
%\leq \frac{n_1}{r}\max_{i \in [r]} \sum_j |\langle \e_i, UU^T \e_j\rangle - \frac{r}{n_1} {\bm 1}_{i=j}|\leq \frac{\mu_1}{\sqrt{r}}.\end{align*}
%
%\noindent Claim now follows by observing that $d\geq r$ and the above equation holds for {\em all} $r$-sized sets $S$.   
%\end{proof}
%Note that the above claim holds with $\delta_d=\mu_1/\sqrt{r},\ \forall d\geq r$. Recall that, Theorem~\ref{thm:main} requires $\delta_d\leq 1/12$. Hence, for $d\geq r\geq 144\mu_1^2$, our assumption in Theorem~\ref{thm:main} is implied by the SIP property of \citep{candes2010power}. 

In the context of universal recovery, a natural question is if any additional assumption is required or the standard $A1$ assumption alone  suffices. Here, we answer this question in negative. Specifically,  we show that if $M$ satisfies  $A1$ only, then universal recovery guarantee for even rank-$2$ matrices  cannot be provided  by using as many as $n_1n_2/4$ observations.%, a universal recovery guarantee cannot be provided for {\em rank-$2$} matrices. %we can show that for universal recovery guarantees, where {\em any} rank-$r$ matrix $M$ can be exactly recovered using only a {\em fixed} set of entries $\Omega$, a condition similar to $A2$ is indeed required and only $A1$ does not suffice. In particular,
\begin{claim}\label{cl:sip1}
  Let $\Omega$ be a fixed set of indices and let $P_\Omega$ be the sampling operator as defined in \eqref{eq:pomega}. Let $n_1=n_2=n$ and let $|\Omega|=n^2/4$. Then, there exists a rank-$2$ matrix $M$ that cannot be recovered exactly from $\Po(M)$. \vspace*{-5pt}
\end{claim}
Now, another  question is if we require a property as strong as $A2$ and if just a lower-bound on $\|U^i\|_2, \|V^j\|_2$ is enough for universal recovery. The proof of the above claim (in appendix) shows that even if $\|U^i\|_2, \|V^j\|_2$  are lower-bounded, then also exact recovery is not possible.
%%% Local Variables: 
%%% mode: latex
%%% TeX-master: "deterministic_MC"
%%% End: 

\section{Simulations}\label{sec:simulations}

\begin{figure*}[ht]
%\vskip -0.2in
  \centering
  \begin{tabular}[ht]{ccc}
    \includegraphics[width=.5\textwidth]{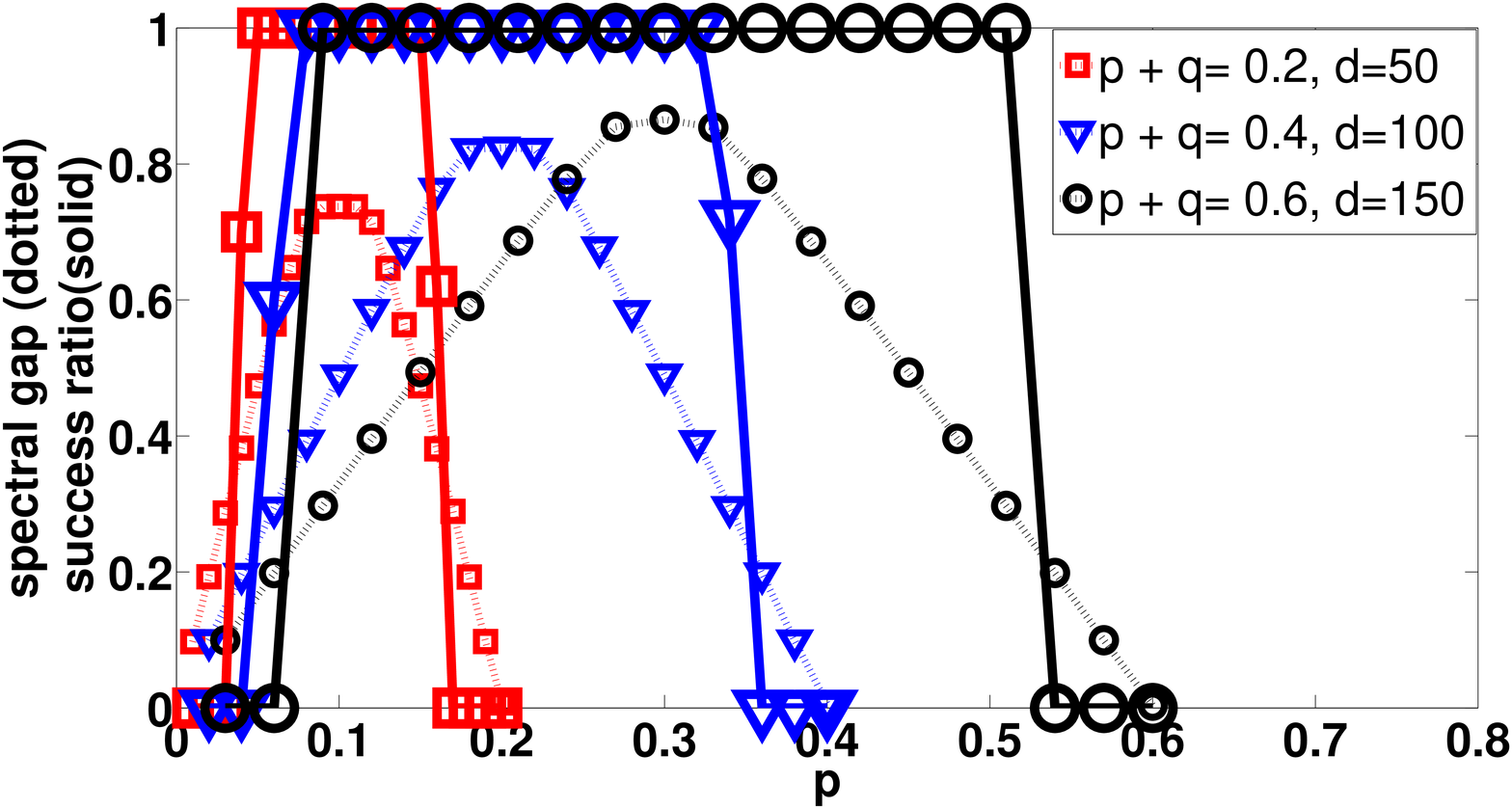}\hspace*{-0pt}&\includegraphics[width=.5\textwidth]{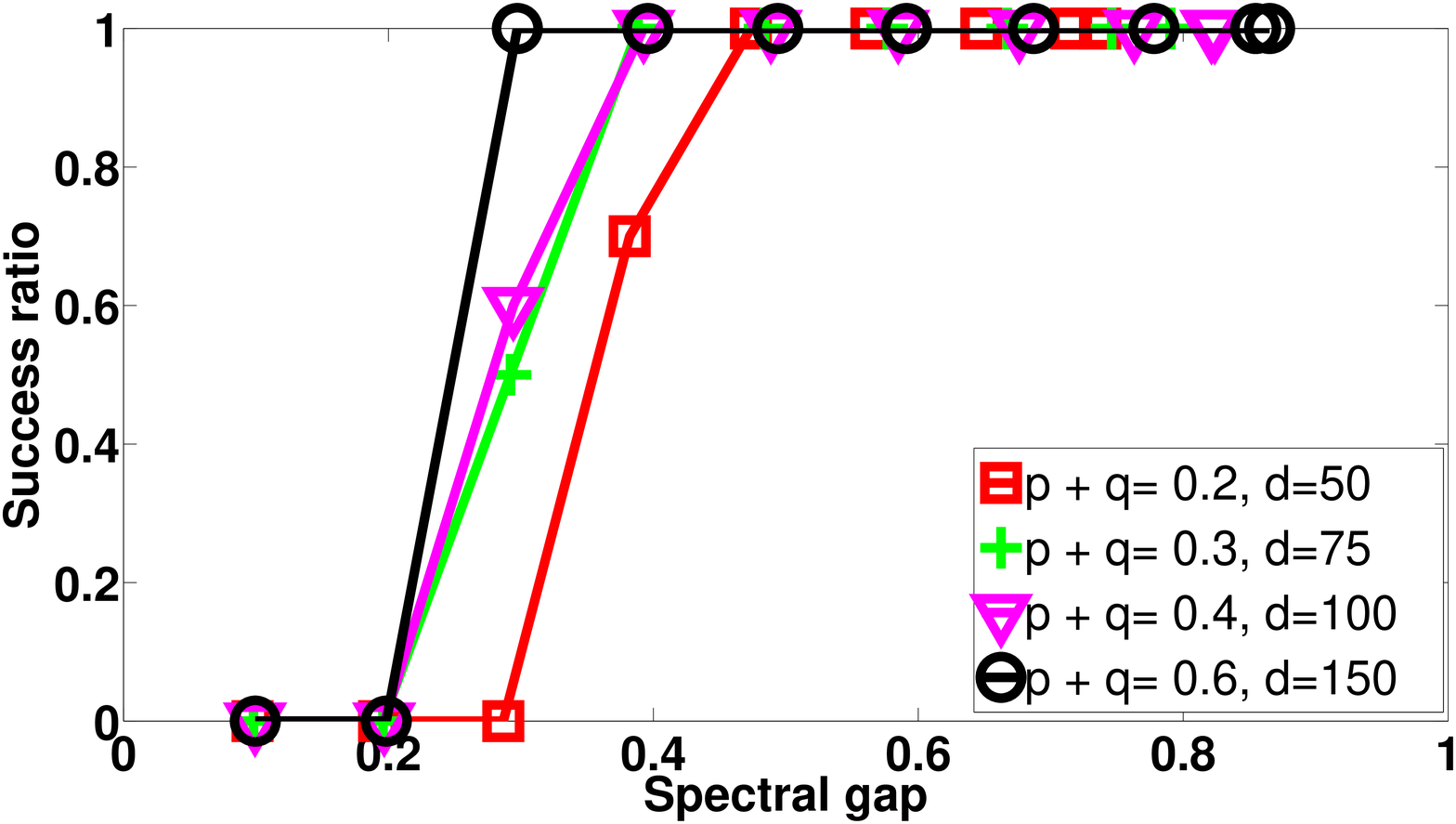}\hspace*{-15pt}\\
{\bf (a)}&{\bf (b)}
  \end{tabular}
  \caption{{\bf(a)}: Figure plots the spectral gap (dotted lines) as well as the the fraction of successful recoveries (solid lines) with varying  $p$, the parameter in stochastic block model. Value of $p+q$ dictates the number of observed entries, i.e., $|\Omega|$.  {\bf (b):} Fraction of successful recovery  vs spectral gap of sampling operator. Clearly, success ratio for matrix recovery show a phase transition type phenomenon w.r.t. the spectral gap. Also, different values of $p+q$, i.e, number of samples do not affect success ratio too much.}
  \label{fig:plot12}
%\vskip -0.2in
\end{figure*} 

\begin{figure*}[ht]
%\vskip 0.2in
  \centering
  \begin{tabular}[ht]{cccc}
    \includegraphics[width=.33\textwidth]{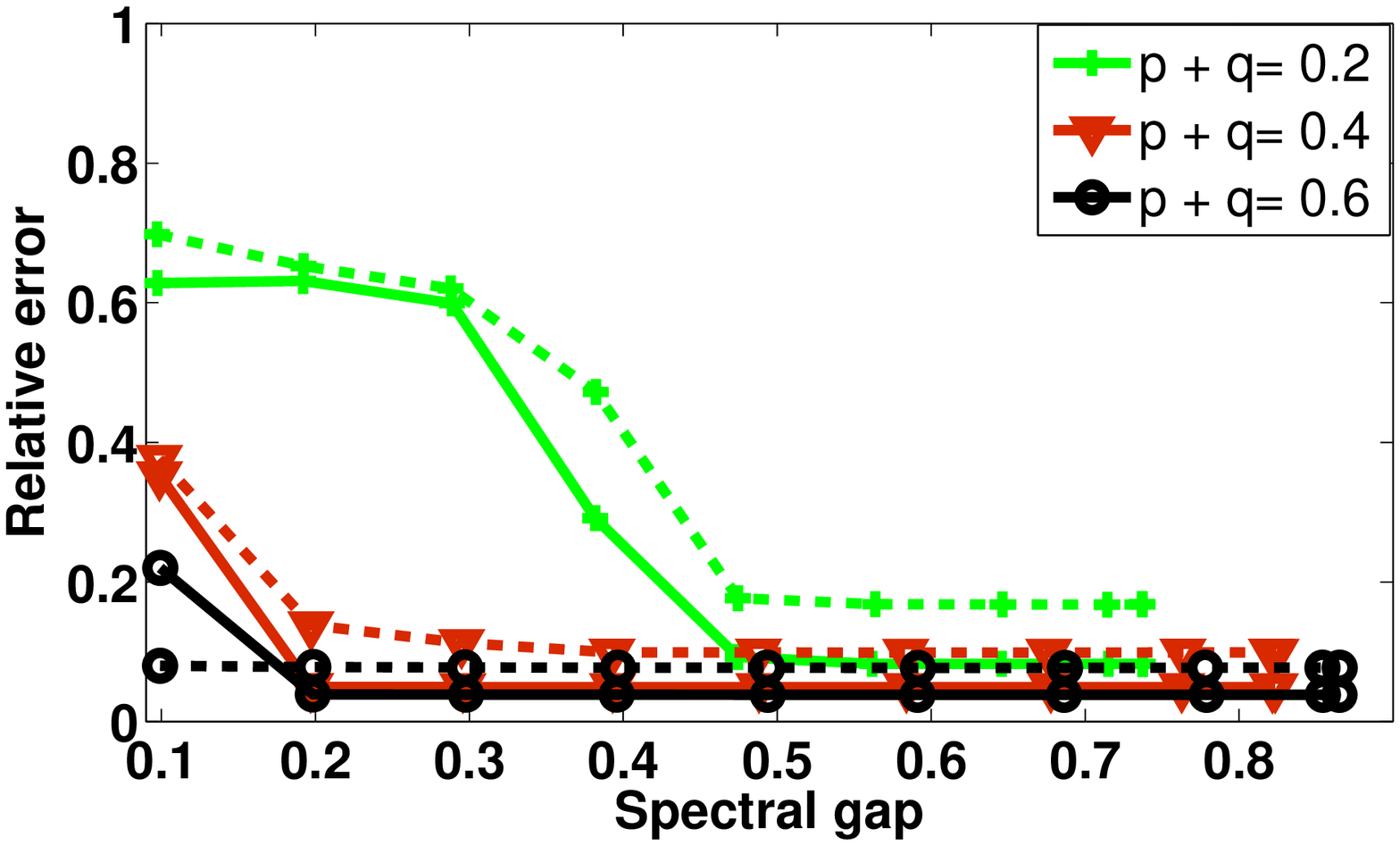}\hspace*{-10pt}&\includegraphics[width=.33\textwidth]{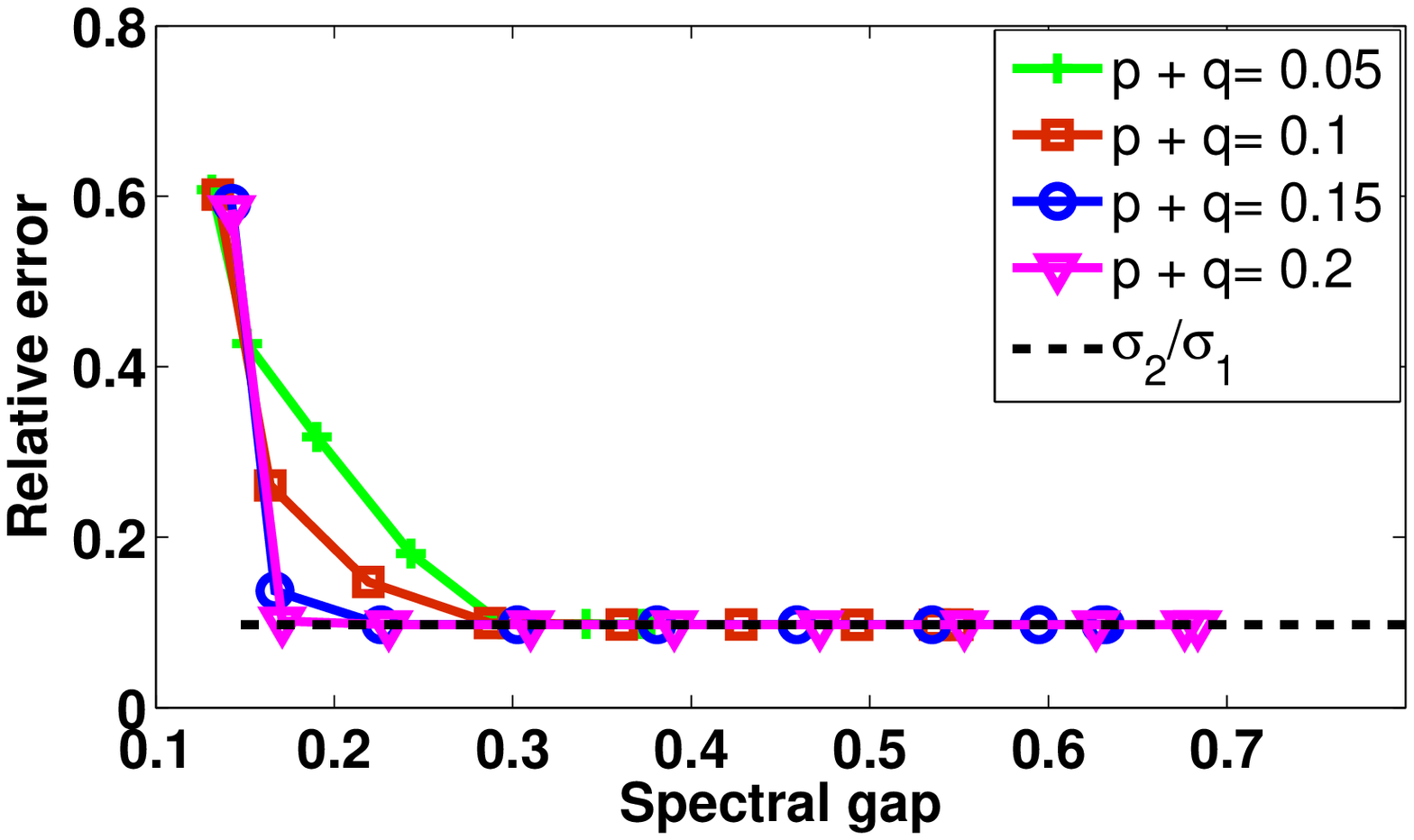}\hspace*{-10pt}&\includegraphics[width=.33\textwidth]{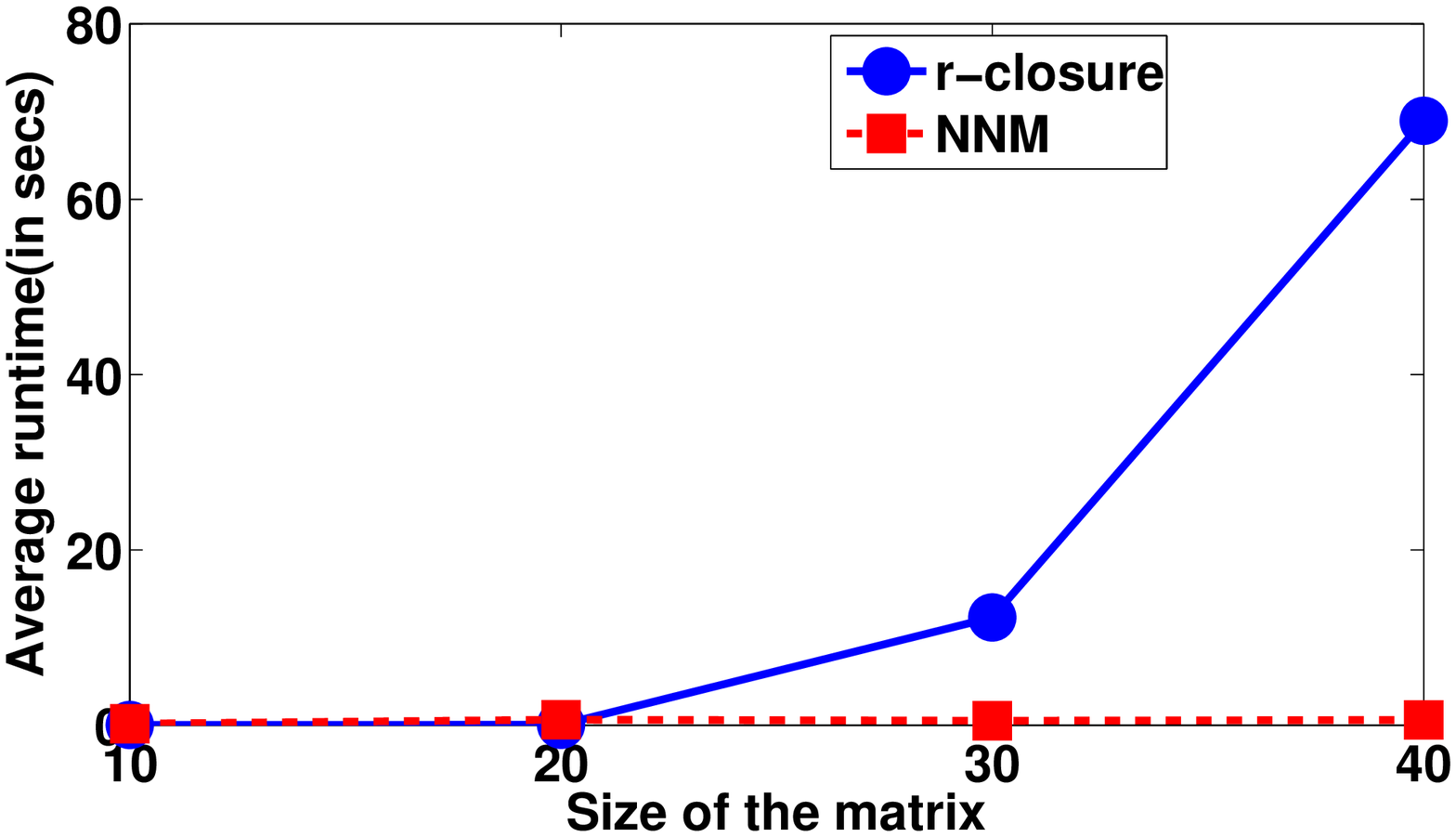}\hspace*{-10pt}\\
{\bf (a)}&{\bf (b)}&{\bf (c)}
  \end{tabular}
  \caption{{\bf(a)}: {\it Noisy samples case:} Relative error in Frobenius norm  vs spectral gap of sampling operator. Perturbed samples of matrix $M+Z$ are observed, where $Z$ is Gaussian noise matrix with $\sigma=\|Z\|_F/ \|M\|_F$. Solid lines correspond to $\sigma=0.1$ and dotted for $\sigma=0.2$.	{\bf (b):} {\it Temperature prediction:} Figure plots relative error in spectral norm vs spectral gap of the sampling operator for different number of samples$(p+q)$. {\bf (c):} {\it Comparison with r-closure algorithm:} Figure compares running time of r-closure algorithm with nuclear norm minimization algorithm.}
  \label{fig:plot34}
%\vskip -0.2in
\end{figure*} 

In this section, we  will present a few  empirical results  on both synthetic and real data sets. The goal of this section is to demonstrate effect of the spectral gap of the sampling graph $\mathcal{G}$ (associated with $\Omega$) on  successful recovery of a matrix. 

First, we use synthetic data sets generated in the following manner. We first sample $U, V \in \R^{500\times 10}$ using standard normal distribution. We then generate rank-$10$ matrix $M$, using $M=UV^T$. %We first generate rank-10 matrices $M\in \R^{500\times 500}$. Let $M=UV^T$, then each entry of $U$ and $V$ are sampled from standard normal distribution. 
As $U, V$ are sampled from the normal distribution, hence w.h.p., $M$ satisfies incoherence assumptions $A1, A2$ mentioned in Section~\ref{sec:formulation}. Next, we generate a sequence of sampling operators $P_\Omega$ (and the associated graph $\mathcal{G}$) with varying (relative) spectral gap($1- \sigma_2(G)/\sigma_1(G)$) by using a stochastic block model. In the basic stochastic block model, the nodes can be thought of as being divided into two clusters. Now, each intra-cluster edge is sampled uniformly with probability $p$ and an inter-cluster edge is sampled uniformly with probability $q$. Note that, when $p=q$, then the spectral gap is largest and when $p=1, q=0$, then spectral gap is smaller as there are two distinct clusters in that case~\citep{nadakuditi2012graph}. %In general, as $p$ deviates more from $q$, the clusters start forming and the spectral gap goes down (see Figure~\ref{fig:plot1}). 

% the probability of each intracluster edge is $p$ and the  i.e, each  of rank the matrices areof rank 10. Matrix $M= UV^T$, where $U$ and $V$ are $500 \times 10$ random matrices with each entry from a standard normal distribution. This ensures that the matrix is incoherent. To create a sequence of sampling operators with varying spectral gap($1- \sigma_2/\sigma_1$) we consider the stochastic block model. When both intracluster and intercluster probabilities ,$p$ and $q$, are same then the graph $G$ has the highest spectral gap. When the probabilities are different then clusters start forming and the spectral gap goes down. 

Note that number of samples generated in this model depends only on the value of $p+ q$. To generate $\Omega$ (i.e., $G$), we first fix a value for $p+ q$, hence fixing the number of samples, and then vary $p, q$, which gives graphs of different spectral gap. As value of $p$ goes from 0 to $\frac{p+ q}{2}$, the spectral gap goes up and from $\frac{p+ q}{2}$ to $p+ q$, the spectral gap goes down.  Figure~\ref{fig:plot12}(a) clearly demonstrates this trend. 

We use an Augmented Lagrangian Method (ALM) based method~\citep{inexactalm} to solve the nuclear norm minimization~\eqref{eq:nnm} problem. % to recover the underlying matrix. We use for Problem~\ref{eq:nnm}. %We solve the nuclear norm minimization~\eqref{eq:nnm} program  using an Augmented Lagrangian Method (ALM) based solver~\citep{inexactalm}. 
A trial is considered to be successful if the relative error (in Frobenius norm) is less than $0.01$. We average over 50 such trials to determine the success ratio.

Figure~\ref{fig:plot12}(a) plots the (relative) spectral gap (dotted lines) and the success ratio (solid lines)  as $p$ varies. Lines of different colors indicate different number of samples ($p+ q$). As expected, the  spectral gap increases initially, as $p$ varies from $0$ to $\frac{p+ q}{2}$ and then it decreases. Moreover, the trend of successful recovery also follows a similar trajectory and hence, is more or less independent of the number of samples (given a particular spectral gap). %One can see that the successful recovery of the matrices depends upon the spectral gap of the sampling operator. 

Figure~\ref{fig:plot12}(b) shows fraction of successful recoveries as the spectral gap increases. Here again, lines of different colors indicate different number of samples ($p+ q$). Clearly, success ratio is positively correlated with the spectral gap of the sampling operator and in fact exhibits a phase transition type of phenomenon. We expect the difference in success ratio for different $p+q$ values to decrease with increasing problem size, i.e, dimensionality of $M$. %Also as the number of samples increase, the curves move left indicating less spectral gap requirement with more samples.

Now, we conduct an experiment to show that spectral gap of $G$ helps in reducing effect of noise as well. To this end, we generated noisy input matrix $(M +Z)$, where $Z$ is a random Gaussian matrix and let $\sigma= ||Z||_F/||M||_F$. We consider two values of $\sigma$, i.e., $\sigma=0.1$ and $0.2$. Figure~\ref{fig:plot34}(a) plots the error(in Frobenius norm) in the recovered matrix against the relative spectral gap in the noisy setting. %As before, lines of different colors correspond to different number of samples ($p+ q$).
Solid lines represent $\sigma=0.1$ and dotted lines represent $\sigma=0.2$. Clearly, larger spectral gap leads to smaller error in recovery. Moreover,  the ``matrix completion denoising'' effect \citep{candes2010matrix} can also be observed. For example, when $||Z||_F=0.1$ and $p+q=0.4$, output error is less than 0.05, and when $||Z||_F=0.2$, error is less than $0.1$ %. As before lines of different colors correspond to different number of samples ($p+ q$). 

%Note that the error of recovered matrix is less than error of input matrix $||Z||_F$ (for example when $||Z||_F=0.1$ and $p+q=0.4$, output error is less than 0.05, and when $||Z||_F=0.2$, error is less than $0.1$) showing the denoising effect of matrix completion, also observed in matrix completion papers with uniform sampling(for ex ~\citep{candes2010matrix}).

%Figure~\ref{fig:plot3} plots the error(in Frobenius norm) in the recovered matrix against the relative spectral gap  in the noisy setting. The input is perturbed by noise $(M +Z)$, where $Z$ is a random Gaussian matrix. Define $\sigma= ||Z||_F/||M||_F$. We consider two values of $\sigma$, 0.1(solid lines) and 0.2(dotted lines). As before lines of different colors correspond to different number of samples ($p+ q$). Note that the error of recovered matrix is less than error of input matrix $||Z||_F$ (for example when $||Z||_F=0.1$ and $p+q=0.4$, output error is less than 0.05, and when $||Z||_F=0.2$, error is less than $0.1$) showing the denoising effect of matrix completion, also observed in matrix completion papers with uniform sampling(for ex ~\citep{candes2010matrix}).

{\bf Temperature prediction:} Finally we take a real dataset of temperature values$(T)$ for 365 days at 316 different locations from~\citep{tempdata}, which has been used to test matrix completion algorithms~\citep{candes2010matrix} before. Note that $T$ is approximately rank-1 matrix with $\sigma_1(T)/||T||_F=0.98$($\sigma_i(T)$ are singular values of $T$). We use the block model sampling scheme to sample entries from $T$, and let the output of~\eqref{eq:nnm} be $\hat{T}$. In figure~\ref{fig:plot34}(b) we plot the error $||\hat{T}-T||/||T||$ for different values of spectral gap of $G$ and for different number of samples $(p+q)$. Note that $||X-T||/||T|| \geq \sigma_2(T)/\sigma_1(T)$ for any rank-1 matrix $X$, and we see that for large enough spectral gap we achieve this bound.

Finally in figure~\ref{fig:plot34}(c) we compare running times of the r-closure algorithm proposed in~\citep{KiralyT12} with the nuclear norm minimization algorithm. While it is noted that this algorithm has better error guarantees, it is combinatorial and takes exponential time to compute.
%%% Local Variables: 
%%% mode: latex
%%% TeX-master: "deterministic_MC"
%%% End: 

\section{Proof of Theorem~\ref{thm:main}}\label{sec:proof_sketch}
In this section, we present the proof of our main result (Theorem~\ref{thm:main}). The main steps of our proof are similar to the proof given by \citep{recht2009simpler}. The main difference is that  the bounds in the existing proof  assume that $\Omega$ is independent of $M$ and hence is not adversarial and holds with high probability. In contrast, for our proofs, bounds are deterministic and our proofs have to work under the assumption that $M$ is adversarially selected for a given $\Omega$. %all our bounds are completely deterministic unlike the ones given in \citep{recht2009simpler} based on concentration results. 

%Similar to \citep{recht2009simpler}, 
The key steps in the proof are: a) provide conditions that an optimal dual solution (or dual certificate) of problem~\eqref{eq:nnm} should satisfy, so that the true matrix $M$ is the {\em unique} optimum of \eqref{eq:nnm}, b) construct such a dual certificate and hence guarantee that $M$ is the unique optimum of \eqref{eq:nnm}.%The proof follows the general line of first showing the conditions a dual certificate should satisfy for $M$ to be optimum of the convex program. Then showing such a dual certificate exists by constructing it. 

We first introduce a few notations required by our proof. For simplicity, we assume that $n_1=n_2=n$. Note that, our proof easily generalizes to case when $n_1\neq n_2$. %Now it is known that \citep{candes2009exact}, $$\partial \|M\|_*=\{UV^T+W,\ s.t.,\ W\in T^\perp,\ \|W\| \leq 1\},$$ where $U, V$ are the singular vectors of $M$. 
Define $T$ which is a subspace of $\R^{n\times n}$, and is span of all matrices of form $UX^T$ and $YV^T$, i.e. all matrices with either row space in $V$ or column space in $U$. Hence, the projection operator $\Pt$ is defined as follows: \begin{align*} \Pt(Z) =UU^T Z + ZVV^T- UU^T ZVV^T = UU^T Z + (I- UU^T)Z VV^T.\end{align*} Hence any matrix in $T$ can be written as $UX^T+YV^T$, for some $X$ and $Y$ such that $Y$ and $U$ are orthogonal to each other. Similarly, we can define the projection operator onto $T^\perp$, the orthogonal complement of $T$: 
$$\Ptp(Z)=(I-UU^T)Z(I-VV^T).$$ 
Now, before presenting conditions on the dual certificate and construction of the dual certificate, we provide a few structural lemmas that show ``goodness'' of operators $\Pt$ and $\Po$. We would like to stress that the key differences of our proof from that of \citep{recht2009simpler} is in fact proofs of these structural lemmas and also in the way we apply Lemma~\ref{lem:pt_infty}. We specifically show (using Lemma~\ref{lem:pt_infty})  that each matrix in the series used in the construction of dual certificate $Y$(discussed later in the section) is incoherent and has small infinity norm. %and also in the way we apply Lemma~\ref{lem:pt_infty}. 

The first lemma proves injectivity of operator $P_\Omega$ on the subspace $T$: 
\begin{lemma}\label{lem:pt}
Let $M=U\Sigma V^T$ satisfies $A1, A2$ and let the graph $\mathcal{G}$ that generates $\Omega$ satisfies $G1, G2$ (see Section~\ref{sec:formulation}). Then, for any matrix $Z \in T$,\vspace*{-5pt} \[\|\fnd \Pt \Po (Z)-Z\|_F \leq \sqrt{2(\delta_d^2+\frac{C^2\mu_0^2 r^2}{d})} \|Z\|_F.\]\vspace*{-5pt}
\end{lemma}
%\begin{proof}
%Here, we provide a high-level proof sketch of the lemma. See appendix for a detailed proof. As, $Z\in T$, hence $Z= UX^T +YV^T$, such that $Y$ and $U$ are orthogonal. Now, using triangular inequality: \begin{multline}\|\fnd\Pt \Po (Z)-Z\|_F \leq \|\fnd\Pt \Po (UX^T) - UX^T\|_F \\+\|\fnd\Pt \Po (YV^T)-YV^T\|_F.\end{multline}
%We consider the first term in the RHS above, the second term is bounded in a similar manner. As $UU^TH$ and $(I-UU^T)H$ are orthogonal for any $H$, hence: 
%\begin{multline*}\hspace*{-15pt}\|\fnd \Pt \Po (UX^T)-UX^T\|_F^2 =\| UU^T(\fnd\Po (UX^T)-UX^T)\|_F^2  \\+ \|(I-UU^T) \fnd\Po (UX^T)VV^T\|_F^2.\end{multline*}
%We bound the first term in RHS above using assumption $A2$, while the second term in RHS is bounded using the spectral gap assumption $G2$ and by using incoherence assumption $A1$. Proof then follows by combining the above two terms. See appendix for the details. 
%\end{proof}

Next, we provide a lemma that characterizes the ``difference'' between $P_\Omega(Z)$ and $Z$, for any incoherent-type $Z \in T$:
\begin{lemma}\label{lem:po_spectral}
Let $Z \in T$, i.e., $Z= UX^T + YV^T$ and $Y$ is orthogonal to $U$, and $X$ and $Y$ be incoherent, i.e., $$\|X^i\|^2 \leq \frac{c_1^2 \mu_0 r}{n},\ \  \|Y^j\|^2 \leq \frac{c_2^2 \mu_0 r}{n}.$$ Let $\Omega$ satisfy the assumptions $ G1$ and $ G2$, then:  \[ \| \fnd \Po(Z)-Z\| \leq (c_1 +c_2)\frac{C\mu_0 r}{\sqrt{d}}.\]
\end{lemma}
The next lemma is a stronger version of Lemma~\ref{lem:pt} for special incoherent-type matrices  $Z\in T$. % $\tilde{Z}=Z-\frac{n}{d}\Pt\Po(Z)$ has small $L_\infty$ norm and furthermore $\tilde{Z}$ is also nearly incoherent.  
%%%%%%%%%%%%%%%%%%%%%-W_k infinity norm lemma-%%%%%%%%%%%%%%%%%%%%%%%%
\begin{lemma}\label{lem:pt_infty}
Let $Z \in T$, i.e., $Z= UX^T + YV^T$ and $Y$ is orthogonal to $U$. Let $X$ and $Y$ be incoherent, i.e., $$\|X^i\|^2 \leq \frac{c_1^2 \mu_0 r}{n},\ \  \|Y^j\|^2 \leq \frac{c_2^2 \mu_0 r}{n}.$$ Let $\tilde{Z} = Z- \fnd \Pt \Po(Z)$. Then, the following holds for $M, \Omega$ that satisfy conditions given in Lemma~\ref{lem:pt}: 
\begin{itemize}
\item $\| \tilde{Z} \|_{\infty} \leq \frac{(c_1 +c_2) \mu_0 r}{n}(\delta_d  + \frac{C\mu_0 r }{ \sqrt{d}}).$ 
\item $\tilde{Z} = U \tilde{X}^T + \tilde{Y} V^T$ and $\tilde{X}$ and $\tilde{Y}$ are incoherent. $\|\tilde{X}^i\|^2 \leq \frac{\mu_0 r}{n} \left(\delta_d c_1+ 2 c_2 \frac{C\mu_0 r}{\sqrt{d}}\right)^2$ and $\|\tilde{Y}^j\|^2 \leq \frac{\mu_0 r}{n} ( \delta_d c_2 +(c_1+ c_2)\frac{C\mu_0 r}{\sqrt{d}})^2.$ 
\end{itemize}
%Above, $c_1, c_2>0$ are global constants. 
\end{lemma}
The proof of the above three lemmas is provided in the appendix. %Now we have stated all our main lemmas and are in a position to prove theorem~\ref{thm:main}.

\noindent {\bf Conditions on the dual certificate}: We now present the lemma that characterizes the conditions a dual certificate should satisfy so that $M$ is the unique optimum of \eqref{eq:nnm}:%based on the subgradient of nuclear norm. %Before presenting the lemma
%Now we will present a lemma  
%Before we present the lemma we need a definition. Define a subspace of $\R^{n \times n}$, $T$ that is span of all 
%%%%%%%%%%%%%%%%%%%%%-approximate dual certificate lemma-%%%%%%%%%%%%%%%%%%%%%%%%
\begin{lemma} \label{lem:dc}
Let $M, \Omega$ satisfy $A1, A2$ and $G1, G2$, respectively. Then, $M$ is the unique optimum of \eqref{eq:nnm}, if there exists a $Y\in \R^{n\times n}$ that satisfies the following: %If there exists a matrix $Y \in \R^{n_1 \times n_2}$ s.t.%\vspace*{-5pt}
\begin{itemize}
\item $\Po(Y)=Y$
\item $||\Pt(Y)-UV^T||_F \leq \sqrt{\frac{d}{8n}}$
\item $||\Ptp(Y)|| < \frac{1}{2}$%\vspace*{-5pt}
\end{itemize}
%then $M$ is the unique optimum of~\eqref{eq:nnm}.
\end{lemma}

Having specified the conditions on dual certificate and also the key structural lemma, we are now ready to present the proof of Theorem~\ref{thm:main}. 
%We now provide a construction of the dual certificate that satisfies the conditions given in the above lemma and hence show that $M$ is exactly recovered by \eqref{eq:nnm}. % we  provide a detailed proof Theorem~\ref{thm:main}. Overall, we show 
\begin{proof}[Proof of Theorem~\ref{thm:main}]
We prove the theorem by constructing a dual certificate $Y$  that satisfies conditions in lemma~\ref{lem:dc} and hence guarantee that $M$ is exactly recovered by \eqref{eq:nnm}. %To prove this theorem we construct a dual certificate $Y$ and show that it satisfies conditions in lemma~\ref{lem:dc}. 
Our construction of $Y$ is similar to the golfing scheme based construction given in \citep{gross2011recovering, recht2009simpler}. In particular, $Y$ is obtained as the $p$-th term of the series given below: 
%Let $W_0 =UV^T$. 
\[ \label{defineY} W_{k+1}= W_k-\fnd\Pt \Po W_{k}, \quad Y_k= \sum_{i=0}^{k-1} \fnd\Po W_{i},\]
where $W_0=UV^T$. That is, $Y=Y_p$ where $p = \lceil\frac{1}{2}\log_3 (\frac{n}{18 C^2 \mu_0^2 r}) \rceil$. Also, define $\alpha=\frac{C\mu_0 r}{\sqrt{d}}$. As $d\geq 36 C^2 \mu_0^2 r^2$, we get: $\alpha\leq \frac{1}{6}$. 

%Now, to prove that the $Y=Y_p$ satisfies conditions of~\ref{lem:dc}, we use results from Lemmas~\ref{lem:pt}, \ref{lem:po_infty} and \ref{lem:pt_infty}. By construction $\Po(Y)=Y$. 
Now, the {\em first condition} of Lemma~\ref{lem:dc} is satisfied trivially by construction as $Y_p$ is a sum of $P_\Omega(W_i)$ terms. 

\noindent {\bf Bounding $||\Pt(Y)-UV^T||_F $:}
By construction: 
\begin{equation}\label{eq:main_1}
 \Pt(Y) -UV^T=\sum_{i=0}^{p-1} \fnd \Pt \Po W_{i} -UV^T =-W_p.\end{equation}
Now, note that each $W_k\in T$. Hence, using Lemma~\ref{lem:pt}, \begin{align}\|W_{k+1}\|_F =\|W_k-\fnd\Pt \Po W_{k}\|_F \\ \leq  \sqrt{2(\delta_d^2+\frac{C^2\mu_0^2 r^2}{d})} \|W_k\|_F \leq\frac{1}{3} \|W_k\|_F,\label{eq:main_2}\end{align} where the last inequality follows by using assumption on $\delta_d$ and by using $\alpha\leq 1/6$. Hence, using \eqref{eq:main_1}, \eqref{eq:main_2}, we have: 
\begin{align}\|\Pt(Y)-UV^T\|_F =\|W_p\|_F \leq  \left(\frac{1}{3}\right)^p \|W_0\|_F  \leq \sqrt{\frac{d}{8n}},\end{align} where the last inequality follows by using $p = \lceil\frac{1}{2}\log_3 (\frac{n}{18 C^2 \mu_0^2 r}) \rceil$ and by using $\|W_0\|_F=\sqrt{r}$. 

\noindent {\bf Bounding $\| \Ptp(Y)\|$:}
%Now using Lemma~\ref{lem:po_infty} and \ref{lem:pt_infty} we can bound $\| \Ptp(Y)\|$. 
Recall that, $W_k \in T$. Now, let $W_k=UX_{W_k}^T+Y_{W_k}V^T$ with $Y_{W_k}$ perpendicular to $U$. Moreover, let, $$\|X^i_{W_k}\|\leq \frac{c_1^{W_k}\sqrt{\mu_0 r}}{\sqrt{n}},\ \ \|Y^i_{W_k}\|\leq \frac{c_2^{W_k}\sqrt{\mu_0 r}}{\sqrt{n}}.$$
Note that, for $W_0=UV^T$, $c_1^{W_0}=1$ and $c_2^{W_0}=0$. Hence, using Lemma~\ref{lem:pt_infty}:
%\|W_1\|_{\infty} \leq  \frac{1}{3}\frac{\mu_0 r}{n},\ \ 
$c_1^{W_1} \leq \frac{1}{6},\ \  c_2^{W_1}\leq \frac{1}{6}.$
Similarly, applying Lemma~\ref{lem:pt_infty} $k$-times, we get: 
\begin{equation}\hspace*{-10pt} c_1^{W_k}=c_2^{W_k}\leq  3^{k-1} \frac{1}{6^{k}}.\label{eq:main_3}\end{equation}
%\begin{equation}\hspace*{-10pt}\|W_k\|_{\infty} \leq 3^{k-2} 4 \frac{1}{6^k} \frac{\mu_0 r}{n},\ \ c_1^{W_k}=c_2^{W_k}\leq  3^{k-1} \frac{1}{6^{k}}.\label{eq:main_3}\end{equation}
%First we will use lemma~\ref{lem:pt_infty} to get bounds on $\|W_k\|_{\infty}$. For $UV^T, c_1=1$ and $c_2=0$. Then we get $\|W_1\|_{\infty} \leq  2\alpha\frac{\mu_0 r}{n}$ and $c_1^{W_1} =\alpha, c_2^{W_1}=\alpha$.

%Using above results we get $\|W_2\|_{\infty} \leq 4\alpha^2 \frac{\mu_0 r}{n}$ and $c_1^{W_2} =3 \alpha^2, c_2^{W_2}=3\alpha^2$. Iterating this way we get $\|W_k\|_{\infty} \leq 3^{k-2} 4 \alpha^k \frac{\mu_0 r}{n}$ and $c_1^{W_k} =c_2^{W_k}= 3^{k-1} \alpha^{k}$.
Now, by using construction of $Y$, by triangle inequality, and by using the fact that $\Ptp$ is a contraction operator: 
%\begin{align}
%\hspace*{-10pt}\| \Ptp(Y)\| \leq %\sum_{k=1}^{p}\|  \fnd \Ptp \Po(W_{k-1}) \|  \\\leq 
%\sum_{k=1}^{p}\|  \fnd  \Po(W_{k-1}) -W_{k-1} \| \leq \sum_{k=1}^{p}  \frac{3Cn}{\sqrt{d}} \|W_{k-1} \|_{\infty}, \label{eq:main_4}\end{align}
\begin{align}
\hspace*{-10pt}\| \Ptp(Y)\| \leq \sum_{k=1}^{p}\|  \fnd  \Po(W_{k-1}) -W_{k-1} \| \leq \sum_{k=1}^{p} (c_1^{W_{k-1}}+ c_2^{W_{k-1}})\frac{C \mu_0 r}{\sqrt{d}} , \label{eq:main_4}\end{align}
where the last inequality follows by using Lemma~\ref{lem:po_spectral}. Now, using \eqref{eq:main_3}, \eqref{eq:main_4}, and Lemma~\ref{lem:pt_infty}, we have:
\begin{align*}
\| \Ptp(Y)\|\leq  \frac{C\mu_0 r}{\sqrt{d}}(1 + 2\cdot\frac{1}{6}(\sum_{k=2}^{p-1} 3^{k-2} \frac{1}{6^{k-2}}))  \leq \alpha(1 + 2\cdot\frac{1}{6}\cdot\frac{1}{1-\frac{1}{2}})  < 1/2.
\end{align*}%for $\alpha \leq \frac{1}{12}.$ The third inequality above follows from lemma~\ref{lem:po_infty}.
Hence, proved. 
\end{proof}
%%% Local Variables: 
%%% mode: latex
%%% TeX-master: "deterministic_MC"
%%% End: 

\section{Conclusions}
In this paper, we provided the first (to the best of our knowledge) universal recovery guarantee for matrix completion. The main observation of the paper is that the spectral gap of $\mathcal{G}$ (that generates $\Omega$) is the key property  that governs recoverability of $M$ using $P_\Omega(M)$ alone.  %is the spectral gap  of the  bipartite graph $G$ that generates $\Omega$. 
For example, if $\mathcal{G}$ is a Ramanujan expander (i.e., $\sigma_2(G)= O(\sqrt{d})$), then % then In particular, if the second largest singular value of the $d$-regular sampling graph $G$ is at most $\sqrt{d}$, 
we have universal recovery guarantees for matrices with strong incoherence property. 

For uniformly sampled $\Omega$, our main result implies exact recovery of constant rank matrices using  $O(n)$ entries, in contrast to the $O(n\log n)$ entries required by the existing analyses. One caveat is that we require stronger incoherence property to obtain the above given sample complexity. %sampledthere exists several interesting randomized as well as deterministic constructions for  Ramanujan graphs, such as, random $d$-regular  graphs. %, deterministic constructions for Ramanujan graphs etc. 
Our results also provide a recipe to determine if a given index set $\Omega$ is enough to recover a low-rank matrix. That is, given  $\Omega$ and its associated graph $\mathcal{G}$, we can measure the spectral gap of $G$ and if it is large enough then our results guarantee exact recovery of strongly incoherent matrices. 

In Section~\ref{sec:conds}, we showed that the standard incoherence assumption alone is not enough for universal recovery and a property similar to $A2$ (see Section~\ref{sec:formulation}) is required. However, it is an open problem to obtain precise information theoretic limits on $\delta_d$ (see $A2$) for universal recovery guarantees. Another interesting research direction is to study the  alternating minimization method  under assumptions given in Section~\ref{sec:formulation}. Finally, we also plan to apply our universal recovery guarantees to specific applications in the domains of sketching and signal-processing. 
%  is large $\Omega$ is sampled according to a bi-partite graph 
%\input{results_altmin}
%
% The following two commands are all you need in the
% initial runs of your .tex file to
% produce the bibliography for the citations in your paper.
%\clearpage
%\newpage
%\bibliographystyle{icml2014}
%\bibliography{reference}%\balancecolumns

%\clearpage
%\newpage
\appendix
\onecolumn
\section{Proof of second part of Theorem~\ref{thm:init}}\label{thm:init2}
\begin{proof}
From the first part of Theorem~\ref{thm:init} we get that $\lV\frac{n}{d}\Po(M)-M\rV \leq  \frac{ C \mu_0 r} {\sqrt{d}} ||M||$. Hence by Weyl\rq{}s inequalities we get that \begin{align*} \lv \fnd\sigma_k( \Po(M)) -\sigma_k(M) \rv \leq \frac{ C \mu_0 r} {\sqrt{d}} ||M||, \end{align*} for all $i$. Since $M$ is rank-$r$ matrix, for any $k \geq r+1$, $\fnd \sigma_k( \Po(M)) \leq  \frac{ C \mu_0 r} {\sqrt{d}} ||M||$. Hence by triangle inequality we get for any $k\geq r$, \begin{align*} \lV \frac{n}{d}P_k(P_\Omega(M))-M\rV &\leq \lV \frac{n}{d}P_k(P_\Omega(M))-\fnd \Po(M)\rV + \lV \fnd \Po(M) -M\rV \\
&\leq \fnd \sigma_{k+1}(P_\Omega(M))  + \lV \fnd \Po(M) -M\rV \\
&\leq  \frac{2 C \mu_0 r} {\sqrt{d}} ||M||. \end{align*}
\end{proof}
\section{Proof of Claim~\ref{cl:sip}}
\begin{proof}
Let $S$ be a set of size $|S|=d$. Since  $\frac{n_1}{d}\sum_{k\in S} U^k{U^k}^T-I$ is a Hermitian matrix,\begin{align}\label{eq:claim11}\lV \frac{n_1}{d}\sum_{k\in S} U^k{U^k}^T-I\rV = \lV \frac{n_1}{d}U_S\rV^2 -1,\end{align} where $U_S$ is a matrix whose columns are $U^k$, $k \in S$. Now we will use equation~\eqref{eq:sip} to bound $\lV U_S \rV$.
\begin{align}\label{eq:claim12}
\|U_S\|^2 &= \max_{x: \|x\|=1} \|U_S x\|^2 = \max_{x: \|x\|=1}\sum_{i, j=1}^d \ip{U^i}{U^j} x_i x_j \nonumber\\
&= \max_{x: \|x\|=1} \sum_{i=1}^d \|U^i\|^2 x_i^2 + \sum_{i \neq j} \ip{U^i}{U^j}x_i x_j \nonumber\\ &\stackrel{\zeta_1}{\leq} \max_{x: \|x\|=1} \|x\|^2 \frac{r+\mu_1 \sqrt{r}}{n} + (d-1)\|x\|^2 \frac{\mu_1 \sqrt{r}}{n} \nonumber\\ &=\frac{r+ d\mu_1 \sqrt{r}}{n}.
\end{align}
$\zeta_1$ follows from~\eqref{eq:sip}. Hence from~\eqref{eq:claim11} and \eqref{eq:claim12} we get, \[ \delta_d \leq \frac{r}{d} + \mu_1\sqrt{r} -1 \leq \mu_1\sqrt{r}.\]
\end{proof}
\section{Proof of Claim~\ref{cl:sip1}}
\begin{proof}
Let $M=UV^T$ where $U\in \R^{n\times 2}$ and $V\in \R^{n\times 2}$ are both orthonormal matrices. Now, let $S=\{j\ s.t.,\ (1,j) \in \Omega \text{ or } (2,j)\in \Omega\}$ %\text{ for any }1\leq i\leq 2 \}$ 
be the set of all the columns of $M$ that have an observed entry in any of the first two rows. 

As $|\Omega|=n^2/4$, hence wlog we can assume that $|S|\leq n/2$. Let $S'=S\cup S_1$, where $S_1$ is any set of columns s.t. $|S'|=n/2$.  Now, construct $U, V$ as follows: 
\begin{equation}V^j=\begin{cases}[\frac{1}{\sqrt{n}}\ \ \ \frac{1}{\sqrt{n}}],\ & \forall j\in S',\\
[\frac{1}{\sqrt{n}}\ \ \ \frac{-1}{\sqrt{n}}],\ & \forall j\not\in S',\end{cases}\vspace*{-5pt}\end{equation}
\begin{equation}\vspace*{-5pt}U^i=\begin{cases}[\frac{1}{\sqrt{n}}\ \ \ \frac{1}{\sqrt{n}}],\ & \forall\ 3\leq i\leq n/2,\\
[\frac{1}{\sqrt{n}}\ \ \ \frac{-1}{\sqrt{n}}],\ & \forall\ n/2+1\leq i\leq n,\\
[a\ \ -a], &\ \ i=1,\\
[b\ \ -b], &\ \ i=2.\end{cases}\end{equation}
%$V^j=[1/\sqrt{n}\ \ 1/\sqrt{n}], \forall j\in S'$ and $V^j=[1/\sqrt{n}\ \ -1/\sqrt{n}], \forall j\not\in S'$. Moreover, let $U^{i}=[1/\sqrt{n}\ \ 1/\sqrt{n}], \forall 3\leq i\leq n/2$ and $U^{i}=[1/\sqrt{n}\ \ 1/\sqrt{n}], \forall n/2+1\leq i\leq n$. Finally, let $U^{1}=[a -a], U^2=[b -b], U^3=[c -c]$.

Note that by construction, $M_{ij}=0, \forall 1\leq i\leq 2, j\in S'$. That is, the first two rows of $\Po(M)$ are all zeros. Since, $U^1$, $U^2$ participate in only those rows. Hence, even if $V$ is known {\em exactly}, one cannot obtain any information about $a, b$ from the observed entries. Only other constraints on $a, b$ comes from orthonormality of $U$, which reduces to $a^2+b^2=2/n$. Now, without violating incoherence assumptions, we can have {\em multiple} solutions to the above given equation that cannot be distinguished from each other. For example, $a=\frac{1}{\sqrt{2n}}$ and $b=\sqrt{\frac{3}{2n}}$, or vice-versa, i.e., $a=\sqrt{\frac{3}{2n}}$ and $b=\frac{1}{\sqrt{2n}}$. 

Hence, exact recovery is not possible for the above given $M$ for {\em any} $\Omega$ s.t. $|\Omega|\leq n^2/4$. 
\end{proof}

\section{Proofs of Lemmas used to prove Theorem~\ref{thm:main}}% of lemmas~\ref{lem:po_infty} and \ref{lem:pt_infty}} 
\label{App:proofs}
In this section we present the proofs of all the lemmas used to prove theorem~\ref{thm:main}.

%%%%%%%%%%%%%%%%%%%%%-PtP lemma-%%%%%%%%%%%%%%%%%%%%%%%%
\begin{lemma*}[\ref{lem:pt}]
Let $M=U\Sigma V^T$ satisfy $A1, A2$ and let the graph $G$ that generates $\Omega$ satisfy $G1, G2$ (see Section~\ref{sec:formulation}). Then, for any matrix $Z \in T$, \[\|\fnd \Pt \Po (Z)-Z\|_F \leq \sqrt{2(\delta_d^2+\frac{C^2\mu_0^2 r^2}{d})} \|Z\|_F.\]
\end{lemma*}
\begin{proof}[Proof of Lemma~\ref{lem:pt}]
%We will first consider the case where both the terms in $Z$ are of rank 1.
%\noindent {\it Rank 1 case:} Let $Z= ux^T +yv^T$, such that $y$ and $u$ are orthogonal. $\|\fnd\Pt \Po (Z)-Z\|_F \leq \|\fnd\Pt \Po (ux^T)-ux^T\|_F +\|\fnd\Pt \Po (yv^T)-yv^T\|_F.$
%
%\noindent Bounding $\|\fnd\Pt \Po (ux^T)-ux^T\|_F $:
%
%\[\|\fnd\Pt \Po (ux^T)-ux^T\|_F^2=\| uu^T(\fnd \Po (ux^T)-ux^T)\|_F^2 +\|(I-uu^T)\fnd \Po (ux^T)vv^T\|_F^2.\]
%
%\begin{align*}
%1) \| uu^T\left(\fnd \Po (ux^T)-ux^T \right)\|_F^2 &=\| u^T\left(\fnd \Po (ux^T)-ux^T \right)\|_2^2 =\sum_j \left(\fnd \sum_i G_{ij}u_i^2 x_j-x_j \right)^2 \\ 
%&\leq \max \{ (\mu_0-1)^2, (\mu_1-1)^2\} \|x\|^2,
%\end{align*}
%where the last inequality follows from assumption~\eqref{eq:assumption2}.
%
%\begin{align*}
%2) \| (I-uu^T)\Po (ux^T)vv^T\|_F^2 =& (I-uu^T)\Po (ux^T)v\|_2^2 =\max_{\tilde{u}: \|\tilde{u}\|\leq 1 \& \tilde{u}^T u=0} \|\tilde{u}^T\Po (ux^T)v\|^2\\
% =& \| (u.\tilde{u})^T G (x.v)\|_F^2  \leq d \|u.\tilde{u}\|^2 \|x.v\|^2 \leq d(\frac{\mu_0 }{n})^2\|x\|^2.
%\end{align*}
%
%\noindent Hence $\|\fnd\Pt \Po (ux^T)\|_F^2 \leq \mu_0^2 \|x\|^2 (1+\frac{1}{d}).$ Similarly $\|\fnd\Pt \Po (yv^T)\|_F^2 \leq \mu_0^2 \|y\|^2 (1+\frac{1}{d}).$ Hence \begin{equation}\|\fnd\Pt \Po (Z)\|_F \leq \mu_0 \sqrt{1+\frac{1}{d}} \sqrt{2} \|Z\|_F.\end{equation}
 Since $Z \in T$, we can write $Z= UX^T +YV^T$, such that $Y$ and $U$ are orthogonal. $\|\fnd\Pt \Po (Z)-Z\|_F \leq \|\fnd\Pt \Po (UX^T) - UX^T\|_F +\|\fnd\Pt \Po (YV^T)-YV^T\|_F.$
Since both the above summands are similar we will bound the first term and then extend the results to the second one. Now, 
\begin{align*}\|\fnd \Pt \Po (UX^T)-UX^T\|_F^2 =&\| UU^T(\fnd\Po (UX^T)-UX^T)\|_F^2 + \|(I-UU^T) \fnd\Po (UX^T)VV^T\|_F^2,\end{align*} as both the terms on RHS are orthogonal to each other. Next, we bound both of these terms individually.
\begin{eqnarray*}
1) \|UU^T(\fnd\Po (UX^T)-UX^T)\|_F^2 &=&\sum_{ij}\left( U^{i^T}(\fnd \sum_k U^k U^{k^T} G_{kj} - I) X^j\right)^2 \\
&=&\sum_j \sum_i\left( U^{i^T}(\fnd \sum_k U^k U^{k^T} G_{kj} - I) X^j\right)^2 \\
&\stackrel{\zeta_1}{=}& \sum_j \lV(\fnd \sum_k U^k U^{k^T} G_{kj} - I) X^j\rV^2  \stackrel{\zeta_2}{\leq} \sum_j \delta_d^2 \|X^j\|^2 =\delta_d^2 \|X\|_F^2,%\vspace*{-5pt}
\end{eqnarray*}
where $\zeta_2$ follows from the assumption $A2$ and from the fact that $G$ is $d$-regular, and $\zeta_1$ follows by using: \begin{align*} \sum_{i=1}^n (U^{i^T} x)^2 =& \sum_{i=1}^n x^T U^i U^{i^T} x =  x^T \left(\sum_{i=1}^nU^i U^{i^T} \right) x = x^T U^T Ux =\|x\|^2.\end{align*}

\begin{align*}
2) \|(I-UU^T)\fnd\Po (UX^T)VV^T\|_F^2 =&\|(I-UU^T)\fnd\Po (UX^T)V\|_F^2=\sum_{i=1}^r \|(I-UU^T)\fnd\Po (UX^T)V_i\|_2^2 \\
\stackrel{\zeta_1}{=}&\sum_{i=1}^r  \max_{\tilde{u}: \|\tilde{u}\|\leq 1\ \&\ \tilde{u}^T U=0}  \|\tilde{u}^T\fnd\Po (UX^T)V_i\|^2, 
\end{align*}
where $\zeta_1$ follows from the definition of the spectral norm. Now,  we  bound  $\|\tilde{u}^T\fnd\Po (UX^T)V_i\|^2$ over $\{ \tilde{u}: \|\tilde{u}\|\leq 1\ \& \ \tilde{u}^T U=0\}$. Note that $\tilde{u}^T U_k=0$ implies that $\tilde{u}.U_k$ is orthogonal to all ones vector.
\begin{align*}
\sum_{i=1}^r  \|\tilde{u}^T\fnd\Po (\sum_{k=1}^r U_k X_k^T)V_i\|^2 =& \sum_{i=1}^r  \|\fnd\sum_{k=1}^r (U_k . \tilde{u})^T G ( X_k .V_i)\|^2  \stackrel{\zeta_1}{\leq} \sum_{i=1}^r  \frac{n^2C^2}{d} \left( \sum_{k=1}^r \|U_k . \tilde{u}\| \|X_k .V_i\| \right)^2 \\
\leq& \sum_{i=1}^r  \frac{n^2C^2}{d} \left( \sum_{k=1}^r \|U_k . \tilde{u}\|^2 \right) \left( \sum_{k=1}^r \|X_k .V_i\|^2 \right) %\leq \sum_{i=1}^r  \frac{n^2}{d} \|\tilde{u}\|^2 \frac{\mu_0 r}{n} \left( \sum_{k=1}^r \|X_i .V_i\|^2 \right) \\
\stackrel{\zeta_2}{\leq} \frac{C^2\mu_0^2 r^2}{d} \|\tilde{u}\|^2 \|X\|_F^2.
\end{align*}
$\zeta_1$ follows from the assumption $G2$ and $\zeta_2$ from incoherence property $A1$.
Using the above two bounds we get $\|\fnd \Pt \Po (UX^T)-UX^T\|_F^2 \leq \|X\|_F^2  (\delta_d^2+\frac{C^2\mu_0^2 r^2}{d}).$ Similarly $\|\fnd\Pt \Po (YV^T)\|_F^2 \leq \|Y\|_F^2  (\delta_d^2+\frac{C^2\mu_0^2 r^2}{d}).$ Hence \[ \|\fnd\Pt \Po (Z)-Z\|_F \leq \sqrt{2(\delta_d^2+\frac{C^2\mu_0^2 r^2}{d})} \|Z\|_F.\]
\end{proof}

%%%%%%%%%%%%%%%%%%%%%-Po spectral norm lemma-%%%%%%%%%%%%%%%%%%%%%%%%
\begin{lemma*}[\ref{lem:po_spectral}]
Let $Z \in T$, i.e., $Z= UX^T + YV^T$ and $Y$ is orthogonal to $U$, and $X$ and $Y$ be incoherent, i.e., $$\|X^i\|^2 \leq \frac{c_1^2 \mu_0 r}{n},\ \  \|Y^j\|^2 \leq \frac{c_2^2 \mu_0 r}{n}.$$ Let $\Omega$ satisfy the assumptions $ G1$ and $ G2$, then:  \[ \| \fnd \Po(Z)-Z\| \leq (c_1 +c_2)\frac{C\mu_0 r}{\sqrt{d}}.\]
\end{lemma*}

\begin{proof}[Proof of Lemma~\ref{lem:po_spectral}]
Note that $\| \fnd \Po(Z)-Z\| \leq \|\fnd \Po(UX^T)-UX^T\| +\|\fnd \Po(YV^T)-YV^T\|$ by triangle inequality. First we will bound $\|\fnd \Po(UX^T)-UX^T\|$. The proof follows the same line as proof of Theorem~\ref{thm:init}.
\begin{align}\label{eq:po_spectral1}
\|\fnd \Po(UX^T)-UX^T\| &= \max_{\{a, b: \|a\|=1, \|b\|=1\}} a^T(\fnd \Po(UX^T)-UX^T)b \nonumber\\
&= \max_{\{a, b: \|a\|=1, \|b\|=1\}} \sum_{i=1}^r\left( \fnd (a.U_i)^T G (X_i .b) - (a^T U_i) (X_i^T b) \right)
\end{align}
Let $a.U_i = \alpha_i \mat{1} + \beta_i \mat{1}_{\perp}^i$. Then $\alpha_i = \frac{(a^T U_i)}{n}$ and $\beta_i^2 \leq \|a. U_i \|^2$. Hence,
\begin{align}\label{eq:po_spectral2}
a^T(\fnd \Po(UX^T)-UX^T)b =\sum_{i=1}^r\left(  (a^T U_i) (X_i^T b)  + \fnd\beta_i{\mat{1}_{\perp}^i}^T G (X_i .b) - (a^T U_i) (X_i^T b) \right) \nonumber\\
\stackrel{\zeta_1}{\leq} \sum_{i=1}^r \frac{Cn}{\sqrt{d}} \beta_i \|X_i .b\| \stackrel{\zeta_2}{\leq} \frac{Cn}{\sqrt{d}} \sqrt{ \sum_{i=1}^r \beta_i^2} \sqrt{\sum_{i=1}^r \|X_i .b\|^2},
\end{align} where $\zeta_1$ follows from assumption $G2$ and $\zeta_2$ from Cauchy-Schwarz inequality.
Now $$\sum_{i=1}^r \beta_i^2 \leq \sum_{i=1}^r\|a. U_i \|^2 = \sum_{j=1}^n \sum_{i=1}^r a_j^2 U_{ji}^2 \leq \frac{\mu_0 r}{n} \sum_{j=1}^n a_j^2 = \frac{\mu_0 r}{n}.$$ Similarly $\sum_{i=1}^r \|X_i .b\|^2 \leq  \frac{c_1^2 \mu_0 r}{n}$. Hence using \eqref{eq:po_spectral1}, \eqref{eq:po_spectral2} and above two inequalities we get $$ \|\fnd \Po(UX^T)-UX^T\|  \leq \frac{c_1 C \mu_0 r}{\sqrt{d}}.$$ Similarly we can show that $\|\fnd \Po(YV^T)-YV^T\|  \leq \frac{c_2 C \mu_0 r}{\sqrt{d}}.$ Hence the lemma follows from the above two bounds.
\end{proof}
%%%%%%%%%%%%%%%%%%%%%-Po infinity norm lemma-%%%%%%%%%%%%%%%%%%%%%%%%
%\begin{lemma*}[\ref{lem:po_infty}]
%Let $\Omega$ satisfy the assumptions $ G1$ and $ G2$, then $\forall Z\in \R^{n\times n}$:  \[ \| \fnd \Po(Z)-Z\| \leq 3\frac{Cn}{\sqrt{d}}\|Z\|_{\infty}.\]
%\end{lemma*}
%
%\begin{proof}[Proof of Lemma~\ref{lem:po_infty}]
%Using definition of spectral norm, we get: 
%\begin{align*}
%\| \fnd \Po(Z)-Z\| &=\max_{a, b: ||a|| \leq 1 \mbox{ and } ||b|| \leq 1}\sum_{ij} Z_{ij} a_i b_j \left(\fnd G_{ij}-1\right) \\
%&=\max_{a, b: ||a|| \leq 1\  ||b|| \leq 1}\sum_{ij} \|Z\|_{\infty} a_i b_j \left(\fnd G_{ij}-1\right) + \max_{a, b: ||a|| \leq 1,\ ||b|| \leq 1}\sum_{ij} (Z_{ij}-\|Z\|_{\infty}) a_i b_j \left(\fnd G_{ij}-1\right)\\
% &\leq 3 \frac{Cn}{\sqrt{d}}\|Z\|_{\infty},
%\end{align*}
%where the last inequality follows using $\sum_{ij} a_i b_j \left(\fnd G_{ij}-1\right) \leq \frac{Cn}{\sqrt{d}}$ (by assumptions $G1$ and $G2$) and $  0\geq Z_{ij}-\|Z\|_{\infty} \geq -2\|Z\|_{\infty} $.
%\end{proof}
%The following lemma shows that the sampling operator $\Po$ gives good approximation also in the infinity norm for any matrix $Z$. The infinity norm is appropriate to consider as the matrices we deal with are incoherent (see next lemma) and incoherent matrices have small infinity norm.

%%%%%%%%%%%%%%%%%%%%%-W_k infinity norm lemma-%%%%%%%%%%%%%%%%%%%%%%%%
\begin{lemma*}[\ref{lem:pt_infty}]
Let $Z \in T$, i.e., $Z= UX^T + YV^T$ and $Y$ is orthogonal to $U$. Let $X$ and $Y$ be incoherent, i.e., $$\|X^i\|^2 \leq \frac{c_1^2 \mu_0 r}{n},\ \  \|Y^j\|^2 \leq \frac{c_2^2 \mu_0 r}{n}.$$ Let $\tilde{Z} = Z- \fnd \Pt \Po(Z)$. Then, the following holds for all $M, \Omega$ that satisfy the conditions given in Lemma~\ref{lem:pt}: 
\begin{itemize}
\item $\| \tilde{Z} \|_{\infty} \leq \frac{(c_1 +c_2) \mu_0 r}{n}(\delta_d  + \frac{C\mu_0 r }{ \sqrt{d}}).$ 
\item $\tilde{Z} = U \tilde{X}^T + \tilde{Y} V^T$ and $\tilde{X}$ and $\tilde{Y}$ are incoherent. $\|\tilde{X}^i\|^2 \leq \frac{\mu_0 r}{n} \left(\delta_d c_1+ 2 c_2 \frac{C\mu_0 r}{\sqrt{d}}\right)^2$ and $\|\tilde{Y}^j\|^2 \leq \frac{\mu_0 r}{n} ( \delta_d c_2 +(c_1+ c_2)\frac{C\mu_0 r}{\sqrt{d}})^2.$ 
\end{itemize}
%Above, $c_1, c_2>0$ are global constants. 
\end{lemma*}

\begin{proof}[Proof of Lemma~\ref{lem:pt_infty}]
\begin{align*}
{ \tilde{Z}_{ij}} = (Z- \fnd \Pt \Po(Z))_{ij} =& \left(UU^T ( UX^T - \fnd \Po(UX^T)) - (I - UU^T) (\fnd \Po(UX^T)) VV^T\right)_{ij}\\ +& \left(( YV^T - \fnd \Po(YV^T)) VV^T - UU^T \fnd \Po(YV^T) (I-VV)^T \right)_{ij},
\end{align*}
where the last equality follows  by using the definition of $\Pt$ and the fact that $Z=\Pt(Z)$. Now, we  bound the first term in the RHS of the above equation. To do this we individually bound  $(UU^T ( UX^T - \fnd \Po(UX^T)))_{ij}$ and $ ((I - UU^T) (\fnd \Po(UX^T)) VV^T)_{ij} $: 
\begin{align*}
(UU^T ( UX^T - \fnd \Po(UX^T)))_{ij} = U^{i^T}\left( I - \fnd \sum_{k=1}^n U^k U^{k^T}G_{kj} \right) X^j \stackrel{\zeta_1}{\leq} \delta_d \|U^i\| \|X^j\| \stackrel{\zeta_2}{\leq} \frac{\delta_d c_1 \mu_0 r}{n},
\end{align*}
where $\zeta_1$ follows from  $A2$ and $\zeta_2$ from the incoherence property $A1$ and the hypothesis of the lemma. 

Similarly, 
\begin{align*}
 \lv ((I - UU^T) (\fnd \Po(UX^T)) VV^T)_{ij} \rv&= \lv U_{\perp}^{i^T} U_{\perp}^T(\fnd \Po(UX^T)) VV^j \rv = \lv \hat{u} \fnd \sum_{k=1}^r\Po(U_k X_k^T) \hat{v} \rv \\
&= \lv \fnd \sum_{k=1}^r(\hat{u}. U_k)^T G (X_k .\hat{v}) \rv \stackrel{\zeta_1}{\leq}  \frac{Cn}{\sqrt{d}} \sum_{k=1}^r \| \hat{u}. U_k\| \|X_k .\hat{v}\|,  
\end{align*}
where $\hat{u}= U_{\perp} U_{\perp}^i$, $\hat{v}= V V^j$, $\mat{1}^T(\hat{u}. U_k) =0$ and $\zeta_1$ follows from  $G2$. Now note that, \[\sum_{k=1}^r \| \hat{u}. U_k\|^2 =\sum_{k=1}^r \sum_{l=1}^n {\ip{U_{\perp}^i}{U_{\perp}^l}}^2 U_{lk}^2 = \sum_{l=1}^n {\ip{U^i}{U^l}}^2 \|U^l\|^2 \leq (\frac{\mu_0 r}{n})^2.\] Using this we can finish the bound as follows:
\begin{align*}
\frac{Cn}{\sqrt{d}} \sum_{k=1}^r \| \hat{u}. U_k\| \|X_k .\hat{v}\| \stackrel{\zeta_1}{\leq}  \frac{Cn}{\sqrt{d}} \sqrt{\frac{\mu_0 r}{n} \frac{c_1^2 \mu_0 r}{n}} \|U^i\| \|V^j\|  \stackrel{\zeta_2}{\leq} \frac{Cn}{\sqrt{d}} \frac{\mu_0 r}{n} \frac{c_1 \mu_0 r}{n}= \frac{C\mu_0^2 r^2 c_1}{n \sqrt{d}},
\end{align*}
where $\zeta_1$ follows from hypothesis of the lemma and $\zeta_2$ from the incoherence property $A1$.

\noindent Putting the two bounds together we get \[\left(UU^T ( UX^T - \fnd \Po(UX^T)) - (I - UU^T) (\fnd \Po(UX^T)) VV^T\right)_{ij} \leq \frac{c_1 \mu_0 r}{n}(\delta_d  + \frac{C\mu_0 r }{ \sqrt{d}}). \]

\noindent Similarly, \[ \left(( YV^T - \fnd \Po(YV^T)) VV^T - UU^T \fnd \Po(YV^T) (I-VV)^T \right)_{ij} \leq \frac{c_2 \mu_0 r}{n}(\delta_d  + \frac{C\mu_0 r}{ \sqrt{d}}).\]

\noindent Hence each element of $Z$ is bounded by, \[ \tilde{Z}_{ij} \leq \frac{(c_1 +c_2) \mu_0 r}{n}(\delta_d  + \frac{C\mu_0 r }{ \sqrt{d}}). \] 

 Now, %we  bound the norm of rows of $\tilde{X}$ as: 
\[ { (U\tilde{X}^T)_{ij}} = (UU^T \tilde{Z} )_{ij}=(UU^T(Z- \fnd \Pt \Po(Z)))_{ij} =(UU^T ( UX^T - \fnd \Po(UX^T) -\fnd \Po(YV^T)))_{ij}.\] 

\noindent Note that $(UU^T \fnd \Po(YV^T))_{ij} =(UU^T \fnd \Po(YV^T) VV^T + UU^T \fnd \Po(YV^T) (I-VV^T))_{ij}.$ Hence, 
\begin{align*}
&{ \|\tilde{X}^j\|^2} = \sum_{i=1}^n (U\tilde{X}^T)_{ij}^2 \\
=& \sum_{i=1}^n \left(U^{i^T}\left( I - \fnd \sum_{k=1}^n U^k U^{k^T}G_{kj} \right) X^j  -U^{i^T} U^T \fnd \Po\left(\sum_{k=1}^r Y_k V_k^T\right) V V^j - U^{i^T} U^T \fnd \Po\left(\sum_{k=1}^r Y_k V_k^T\right) V_{\perp} V_{\perp}^j \right)^2 \\
\stackrel{\zeta_1}{=}&\lV \left( I - \fnd \sum_{k=1}^n U^k U^{k^T}G_{kj} \right) X^j - U^T \fnd \Po\left(\sum_{k=1}^r Y_k V_k^T\right) V V^j - U^T \fnd \Po\left(\sum_{k=1}^r Y_k V_k^T\right) V_{\perp} V_{\perp}^j \rV^2 \\
\leq&\left(\lV \left( I - \fnd \sum_{k=1}^n U^k U^{k^T}G_{kj} \right) X^j\rV + \lV U^T \fnd \Po\left(\sum_{k=1}^r Y_k V_k^T\right) V V^j\rV + \lV U^T \fnd \Po\left(\sum_{k=1}^r Y_k V_k^T\right) V_{\perp} V_{\perp}^j \rV \right)^2,%\\
%\leq& \left( \delta_d \sqrt{\frac{c_1^2 \mu_0 r}{n}} +  \sqrt{\frac{C^2 c_2^2 \mu_0^3 r^3}{nd}}+  \sqrt{\frac{C^2 c_2^2 \mu_0^3 r^3}{nd}}\right)^2 \leq \frac{\mu_0 r}{n} \left(\delta_d c_1+ 2 c_2 \frac{C \mu_0 r}{\sqrt{d}}\right)^2.
\end{align*}
where, $\zeta_1$ follows by the following: \[ \sum_{i=1}^n (U^{i^T} x)^2 = \sum_{i=1}^n x^T U^i U^{i^T} x =  x^T \left(\sum_{i=1}^nU^i U^{i^T} \right) x = x^T U^T Ux =\|x\|^2.\]
Next, we bound each of the above three terms individually. First term $\lV \left( I - \fnd \sum_{k=1}^n U^k U^{k^T}G_{kj} \right) X^j\rV$ is bounded by $\delta_d \sqrt{\frac{c_1^2 \mu_0 r}{n}}$, which follows from the assumption $A2$ and the hypothesis of the lemma. Next, we consider the second and third terms.

\begin{align*}
2) &\lV U^T \fnd \Po\left(\sum_{k=1}^r Y_k V_k^T\right) V V^j\rV =\max_{a: \|a\| \leq 1} a^T U^T \fnd \Po\left(\sum_{k=1}^r Y_k V_k^T\right) V V^j = \max_{a: \|a\| \leq 1} \fnd \sum_{k=1}^r \left( (Ua . Y_k)^T G (V_k .\hat{v}) \right)\\
\stackrel{\zeta_1}{\leq}& \max_{a: \|a\| \leq 1} \frac{Cn}{\sqrt{d}} \sum_{k=1}^r \|Ua . Y_k \| \|V_k .\hat{v}\| \leq  \max_{a: \|a\| \leq 1} \frac{Cn}{\sqrt{d}} \sqrt{\sum_{k=1}^r  \|Ua . Y_k \|^2}\sqrt{\sum_{k=1}^r \|V_k .\hat{v}\|^2 } 
\stackrel{\zeta_2}{\leq} \max_{a: \|a\| \leq 1} \frac{Cn}{\sqrt{d}} \sqrt{\frac{c_2^2 \mu_0 r}{n}} \|a\|  \frac{ \mu_0 r}{n} \\
{=}& \sqrt{\frac{C^2 c_2^2 \mu_0^3 r^3}{nd}},
\end{align*}
where $\hat{v}= V V^j$ and $a^T U^T Y_k =0$. $\zeta_1$ follows from the assumption $G2$ and $\zeta_2$ from the assumption $A1$ and the hypothesis of the lemma.

\begin{align*}
3) &\lV U^T \fnd \Po\left(\sum_{k=1}^r Y_k V_k^T\right) V_{\perp} V_{\perp}^j\rV =\max_{a: \|a\| \leq 1} a^T U^T \fnd \Po\left(\sum_{k=1}^r Y_k V_k^T\right) V_{\perp} V_{\perp}^j = \max_{a: \|a\| \leq 1} \fnd \sum_{k=1}^r \left( (Ua . Y_k)^T G (V_k .\hat{v}) \right)\\
\stackrel{\zeta_1}{\leq}& \max_{a: \|a\| \leq 1} \frac{Cn}{\sqrt{d}} \sum_{k=1}^r \|Ua . Y_k \| \|V_k .\hat{v}\| \leq  \max_{a: \|a\| \leq 1} \frac{Cn}{\sqrt{d}} \sqrt{\sum_{k=1}^r  \|Ua . Y_k \|^2}\sqrt{\sum_{k=1}^r \|V_k .\hat{v}\|^2 } 
\stackrel{\zeta_2}{\leq}  \max_{a: \|a\| \leq 1} \frac{Cn}{\sqrt{d}} \sqrt{\frac{c_2^2 \mu_0 r}{n}} \|a\|  \frac{ \mu_0 r}{n} \\
{=}& \sqrt{\frac{C^2 c_2^2 \mu_0^3 r^3}{nd}},
\end{align*}
where $\hat{v}= V_{\perp} V_{\perp}^j$ and $a^T U^T Y_k =0$. $\zeta_1$ follows from $G2$ and $\zeta_2$ from  $A1$ and the hypothesis of the lemma. Using all the three bounds we can finally bound $\|\tilde{X}^j\|^2$. \[\|\tilde{X}^j\|^2 \leq \left( \delta_d \sqrt{\frac{c_1^2 \mu_0 r}{n}} +  \sqrt{\frac{C^2 c_2^2 \mu_0^3 r^3}{nd}}+  \sqrt{\frac{C^2 c_2^2 \mu_0^3 r^3}{nd}}\right)^2 = \frac{\mu_0 r}{n} \left(\delta_d c_1+ 2 c_2 \frac{C \mu_0 r}{\sqrt{d}}\right)^2.\]

Now, we  bound the norm of rows of $\tilde{Y}$.
\begin{align*}
{ \|\tilde{Y}^i\|^2} =& \sum_{j=1}^n (\tilde{Y}V^T)_{ij}^2 \\ =& \sum_{j=1}^n \left(Y^{i^T} \left( I- \fnd\sum_{k=1}^n V^k V^{k^T}G_{ik} \right) V^j + U^{i^T} U^T \fnd \Po\left(\sum_{k=1}^r Y_k V_k^T\right) V V^j -  U_{\perp}^{i^T} U_{\perp}^T\left(\fnd \Po(UX^T)\right) VV^j \right)^2 \\
=& \lV Y^{i^T} \left( I- \fnd\sum_{k=1}^n V^k V^{k^T}G_{ik} \right)  + U^{i^T} U^T \fnd \Po\left(\sum_{k=1}^r Y_k V_k^T\right) V  -  U_{\perp}^{i^T} U_{\perp}^T\left(\fnd \Po(UX^T)\right) V\rV^2 \\
\leq& \left(\lV Y^{i^T} \left( I- \fnd\sum_{k=1}^n V^k V^{k^T}G_{ik} \right) \rV + \lV U^{i^T} U^T \fnd \Po\left(\sum_{k=1}^r Y_k V_k^T\right) V \rV + \lV U_{\perp}^{i^T} U_{\perp}^T\left(\fnd \Po(UX^T)\right) V \rV \right)^2 %\\
%\leq & \left( \delta_d \sqrt{\frac{c_2^2 \mu_0 r}{n}} + \sqrt{\frac{C^2 c_2^2 \mu_0^3 r^3}{nd}}+ \sqrt{\frac{C^2 c_1^2 \mu_0^3 r^3}{nd}} \right)^2 \leq \frac{\mu_0 r}{n} \left(\delta_d c_2 + (c_1 +c_2)\frac{C\mu_0 r}{\sqrt{d}}\right)^2.
\end{align*}
Next, we  bound each of the above three terms individually. First term $\lV Y^{i^T} \left( I- \fnd\sum_{k=1}^n V^k V^{k^T}G_{ik} \right) \rV$ is bounded by $\delta_d \sqrt{\frac{c_2^2 \mu_0 r}{n}}$, which follows from  $A2$ and the hypothesis of the lemma. Now, we  bound the second and third terms.

\begin{align*}
2) &\lV U^{i^T} U^T \fnd \Po\left(\sum_{k=1}^r Y_k V_k^T\right) V \rV^2 = \max_{b : \|b\| \leq 1} U^{i^T} U^T \fnd \Po\left(\sum_{k=1}^r Y_k V_k^T\right) Vb =  \max_{b : \|b\| \leq 1} \fnd \sum_{k=1}^r \left( (\hat{u} . Y_k)^T G (V_k . Vb)\right) \\
\stackrel{\zeta_1}{\leq}& \max_{b : \|b\| \leq 1} \frac{Cn}{\sqrt{d}}\sum_{k=1}^r \| \hat{u} . Y_k \| \| V_k . Vb\| \leq  \max_{b : \|b\| \leq 1} \frac{Cn}{\sqrt{d}} \sqrt{\sum_{k=1}^r \| \hat{u} . Y_k \|^2}  \sqrt{\sum_{k=1}^r \| V_k . Vb\|^2 } \stackrel{\zeta_2}{\leq}   \max_{b : \|b\| \leq 1} \frac{Cn}{\sqrt{d}} \frac{c_2 \mu_0 r}{n} \sqrt{\frac{\mu_0 r}{n}} \|b\| \\
{=}& \sqrt{\frac{C^2 c_2^2 \mu_0^3 r^3}{nd}},
\end{align*}
where $\hat{u}= U U^i$ and $\mat{1}^T(\tilde{u}. Y_k) =0$. $\zeta_1$ follows from  $G2$ and $\zeta_2$ from  $A1$ and the hypothesis of the lemma. 

\begin{align*}
3)&\lV U_{\perp}^{i^T} U_{\perp}^T\left(\fnd \Po(UX^T)\right) V \rV = \max_{b : \|b\| \leq 1} U_{\perp}^{i^T} U_{\perp}^T\left(\fnd \Po(UX^T)\right) Vb =  \max_{b : \|b\| \leq 1} \fnd \sum_{k=1}^r \left( (\hat{u} . U_k)^T G (X_k . Vb)\right) \\
\stackrel{\zeta_1}{\leq}& \max_{b : \|b\| \leq 1} \frac{Cn}{\sqrt{d}}\sum_{k=1}^r \| \hat{u} . U_k \| \| X_k . Vb\| \leq \max_{b : \|b\| \leq 1} \frac{Cn}{\sqrt{d}} \sqrt{\sum_{k=1}^r \| \hat{u} . U_k \|^2}  \sqrt{\sum_{k=1}^r \| X_k . Vb\|^2 } \stackrel{\zeta_2}{\leq}   \max_{b : \|b\| \leq 1} \frac{Cn}{\sqrt{d}} \frac{ \mu_0 r}{n} \sqrt{\frac{c_1^2 \mu_0 r}{n}} \|b\| \\
{=}& \sqrt{\frac{C^2 c_1^2 \mu_0^3 r^3}{nd}},
\end{align*}
where $\hat{u}= U_{\perp} U_{\perp}^i$ and $\mat{1}^T(\tilde{u}. U_k) =0$.  $\zeta_1$ follows from  $G2$ and $\zeta_2$ from  $A1$ and the hypothesis of the lemma. %The fifth inequality follows from,
%\[\sum_{k=1}^r \| \hat{u}. U_k\|^2 =\sum_{k=1}^r \sum_{l=1}^n {\ip{U_{\perp}^i}{U_{\perp}^l}}^2 U_{lk}^2 = \sum_{l=1}^n {\ip{U^i}{U^l}}^2 \|U^l\|^2 \leq (\frac{\mu_0 r}{n})^2.\]
Using all the three bounds we can finally bound $\|\tilde{Y}^i\|^2$. \[\|\tilde{Y}^i\|^2 \leq \left( \delta_d \sqrt{\frac{c_2^2 \mu_0 r}{n}} + \sqrt{\frac{C^2 c_2^2 \mu_0^3 r^3}{nd}}+ \sqrt{\frac{C^2 c_1^2 \mu_0^3 r^3}{nd}} \right)^2 = \frac{\mu_0 r}{n} \left(\delta_d c_2 + (c_1 +c_2)\frac{C\mu_0 r}{\sqrt{d}}\right)^2.\]
\end{proof}

%%%%%%%%%%%%%%%%%%%%%-dual certificate lemma-%%%%%%%%%%%%%%%%%%%%%%%%
\begin{lemma*}[\ref{lem:dc}]
Let $M, \Omega$ satisfy $A1, A2$ and $G1, G2$, respectively. Then, $M$ is the unique optimum of \eqref{eq:nnm}, if there exists a $Y\in \R^{n\times n}$ that satisfies the following: %If there exists a matrix $Y \in \R^{n_1 \times n_2}$ s.t.%\vspace*{-5pt}
\begin{itemize}
\item $\Po(Y)=Y$
\item $||\Pt(Y)-UV^T||_F \leq \sqrt{\frac{d}{8n}}$
\item $||\Ptp(Y)|| < \frac{1}{2}$%\vspace*{-5pt}
\end{itemize}
%then $M$ is the unique optimum of~\eqref{eq:nnm}.
\end{lemma*}

\begin{proof}[Proof of Lemma~\ref{lem:dc}]
For any $Z$ such that, $\Po(Z)=0$, implies $\|\Po \Ptp (Z)\| = \|\Po \Pt(Z)\|$. Also let $\delta_d =\frac{C\mu_0 r}{\sqrt{d}} =\alpha$.
\begin{align*}
 \|\Po \Pt(Z)\| _F &=\ip{\Pt(Z)}{ \Po \Pt(Z)}  \stackrel{\zeta_1}{\geq} \frac{d}{n} (1-\sqrt{2(\delta_d^2+\frac{C^2\mu_0^2 r^2}{d})})  ||\Pt(Z)||_F^2 = \frac{d}{n} (1-2\alpha) ||\Pt(Z)||_F^2 \\ &> \frac{d}{2n} ||\Pt(Z)||_F^2,
\end{align*}
for $\alpha < \frac{1}{4}$. $\zeta_1$ follows from Lemma~\ref{lem:pt}. Also note that  $\| \Po \Ptp(Z)\|_F \leq \| \Ptp(Z)\|_F.$ Hence, \[ \|\Ptp(Z)\|_* \geq \|\Ptp(Z)\|_F >\sqrt{\frac{d}{2n}} ||\Pt(Z)||_F.\]
Now choose $U_{\perp}$ and $V_{\perp}$ from the SVD of $\Ptp(Z)$, which ensures that $\ip{U_{\perp}V_{\perp}^T}{\Ptp(Z)}=\|\Ptp(Z)\|_*$.  Now, 
\begin{align*}
\|M+Z\|_*  \stackrel{\zeta_1}{\geq}& \ip{UV^T+ U_{\perp}V_{\perp}^T}{M+Z} \\ =&\|M\|_* + \ip{UV^T+ U_{\perp}V_{\perp}^T}{Z} \\
\stackrel{\zeta_2}{=}&\|M\|_* + \ip{UV^T+ U_{\perp}V_{\perp}^T}{Z} -\ip{Y}{Z} \\
=&\|M\|_* + \ip{UV^T -\Pt(Y)}{\Pt(Z)} +\ip{U_{\perp}V_{\perp}^T- \Ptp(Y)}{\Ptp(Z)} \\
\stackrel{\zeta_3}{\geq}& \|M\|_* -\|UV^T -\Pt(Y)\|_F \|\Pt(Z)\|_F + \|\Ptp(Z)\|_* - \|\Ptp(Y)\| \|\Ptp(Z)\|_* \\
>& \|M\|_* -\|UV^T -\Pt(Y)\|_F\|\Pt(Z)\|_F + (1-\|\Ptp(Y)\|)\sqrt{\frac{d}{2n}} ||\Pt(Z)||_F \\ \stackrel{\zeta_4}{>}& \|M\|_* .
\end{align*}
$\zeta_1$ follows from the Holder's inequality and the fact that $\|UV^T+ U_{\perp}V_{\perp}^T\| =1$; $\zeta_2$ from $\ip{Y}{Z} =\ip{\Po(Y)}{Z}=0$; $\zeta_3$ again from the Holder's inequality; and $\zeta_4$ from the hypothesis of lemma.
\end{proof}

\end{document}